%% file: main.tex
\documentclass{article}

\usepackage{microtype}
\usepackage{graphicx}
\usepackage{zref}
\usepackage{booktabs} 
\usepackage{amsmath}
\usepackage{amssymb}
\usepackage{palatino}
\usepackage{graphicx}
\usepackage{float}
\usepackage{natbib}
\usepackage{subfigure}
\usepackage{tabulary}
\usepackage{hyperref}       
\usepackage{url}            
\usepackage{booktabs}       
\usepackage{amsfonts}       
\usepackage{nicefrac}       
\usepackage{microtype}      
\usepackage{hyperref}
\usepackage{booktabs}
\usepackage{amsfonts}
\usepackage{amstext}
\usepackage{amsthm}
\usepackage{bm}
\usepackage{wrapfig}
\usepackage{comment}
\usepackage{color}
\usepackage{float}
\usepackage{enumitem}
\usepackage{MnSymbol,wasysym,marvosym,ifsym}
\usepackage{algorithm}
\usepackage{algorithmic}
\usepackage[utf8]{inputenc}
\usepackage{caption}
\usepackage{diagbox}

\input{cmd.tex}

\usepackage{hyperref}

\usepackage[accepted]{icml2021}

\icmltitlerunning{
Can Subnetwork Structure be the Key to Out-of-Distribution Generalization?
}

\begin{document}

\twocolumn[
\icmltitle{
Can Subnetwork Structure be the Key to Out-of-Distribution Generalization?
}





\begin{icmlauthorlist}
\icmlauthor{Dinghuai Zhang}{mila}
\icmlauthor{Kartik Ahuja}{mila}
\icmlauthor{Yilun Xu}{mit}
\icmlauthor{Yisen Wang}{pku}
\icmlauthor{Aaron Courville}{mila}
\end{icmlauthorlist}

\icmlaffiliation{mila}{Mila - Quebec AI Institute}
\icmlaffiliation{mit}{CSAIL, Massachusetts Institute of Technology}
\icmlaffiliation{pku}{Key Lab of Machine Perception (MoE), School of EECS, Peking University}

\icmlcorrespondingauthor{Dinghuai Zhang}{dinghuai.zhang@mila.quebec}

\icmlkeywords{Machine Learning, ICML}

\vskip 0.3in
]

\printAffiliationsAndNotice{}  

\newtheorem*{theorem*}{Theorem}
\newtheorem*{remark}{Remark}

\newcommand{\fix}{\marginpar{FIX}}
\newcommand{\new}{\marginpar{NEW}}

\def\viz{\emph{viz}\onedot}

\newcommand{\Todo}[1]{\textcolor{red}{TODO: #1}\\}
\newcommand{\todo}[1]{\textcolor{gray}{TODO: #1}}
\newcommand{\tocite}[1]{\textcolor{gray}{add cite #1}}
\newcommand{\red}[1]{\textcolor{red}{#1}}
\newcommand{\med}[1]{\textcolor{magenta}{#1}}
\newcommand{\blue}[1]{\textcolor{blue}{#1}}
\newcommand{\gray}[1]{\textcolor{gray}{#1}}
\newcommand{\green}[1]{\textcolor{green}{#1}}
\newcommand{\zdh}[1]{\textcolor{gray}{dinghuai: #1}}

\newcommand{\tr}[1]{\text{tr}\left(#1\right)}
\newcommand{\N}{\mathcal{N}}
\newcommand{\Sm}{\Sigma}
\newcommand{\norm}[1]{\left\lVert#1\right\rVert}
\def\viz{\emph{viz}\onedot}
\newcommand{\diag}[1]{\text{diag}\left(#1\right)}
\newcommand{\dt}{\frac{\dif}{\dif t}}

\newcommand{\E}{\mathbb{E}}
\newcommand{\V}{\mathbb{V}}
\global\long\def\R{\mathbb{R}}%
\newcommand{\z}{\boldsymbol{z}}
\renewcommand{\vec}[1]{\boldsymbol{#1}}
\renewcommand{\vv}[1]{\boldsymbol{#1}}
\newcommand{\normal}{\mathcal{N}}
\newcommand{\F}{\mathcal{F}}
\newcommand{\B}{\mathcal{B}}
\newcommand{\D}{\mathbb{D}}

\newcommand{\e}{\mathbf{e}}
\newcommand{\etr}{\mathbf{e}_{tr}}
\newcommand{\x}{\mathbf{x}}
\newcommand{\X}{\mathbf{X}}
\newcommand{\y}{\mathbf{y}}
\newcommand{\xc}{\mathbf{x}_{\text{inv}}}
\newcommand{\Xc}{\mathbf{X}_{\text{inv}}}
\newcommand{\xe}{\mathbf{x}_{e}}
\newcommand{\Xe}{\mathbf{X}_{e}}
\newcommand{\hbb}{\hat{\boldsymbol{\beta}}}
\newcommand{\bc}{{\boldsymbol{\beta}_{\text{inv}}}}
\newcommand{\hbc}{\hat{\boldsymbol{\beta}}_{\text{inv}}}
\newcommand{\be}{{\boldsymbol{\beta}_{e}}}
\newcommand{\hbe}{\hat{\boldsymbol{\beta}}_{e}}
\newcommand{\mmu}{{\boldsymbol{\mu}}}
\newcommand{\muc}{{\boldsymbol{\mu}_{\text{inv}}}}
\newcommand{\mue}{{\boldsymbol{\mu}_{e}}}
\newcommand{\I}{\mathbf{I}}





\begin{abstract}
Can models with particular structure avoid being biased towards spurious correlation in out-of-distribution (OOD) generalization? \citet{peters2016causal} provides a positive answer for linear cases.
In this paper, we use a functional modular probing method to 
analyze deep model structures under OOD setting. 
We demonstrate that even in biased models (which focus on spurious correlation) there still exist unbiased functional subnetworks. 
Furthermore, we articulate and demonstrate the functional lottery ticket hypothesis: full network contains a subnetwork that can achieve better OOD performance.
We then propose Modular Risk Minimization to solve the subnetwork selection problem. Our algorithm learns the subnetwork structure from a given dataset, and can be combined with any other OOD regularization methods. Experiments on various OOD generalization tasks corroborate the effectiveness of our method.  
\end{abstract}

\section{Introduction}

Despite the remarkable progress we have witnessed in neural-network-based machine learning, the stories of failures continue to accumulate \cite{geirhos2020shortcut}.  Many of these failures are attributed to models exploiting spurious correlations or shortcuts (i.e. factors that are not used to generate the label).  A colloquial example comes from \cite{beery2018recognition} where the authors show how a neural network trained to distinguish cows from camels exploits shortcut such as background color for prediction. In a much more concerning example, \cite{degrave2020ai} show how machine learning systems trained to detect COVID-19 exploited the data source (e.g., hospital) to artificially boost inference performance.

\begin{figure}[t]
    \centering
    \includegraphics[width=0.6\columnwidth]{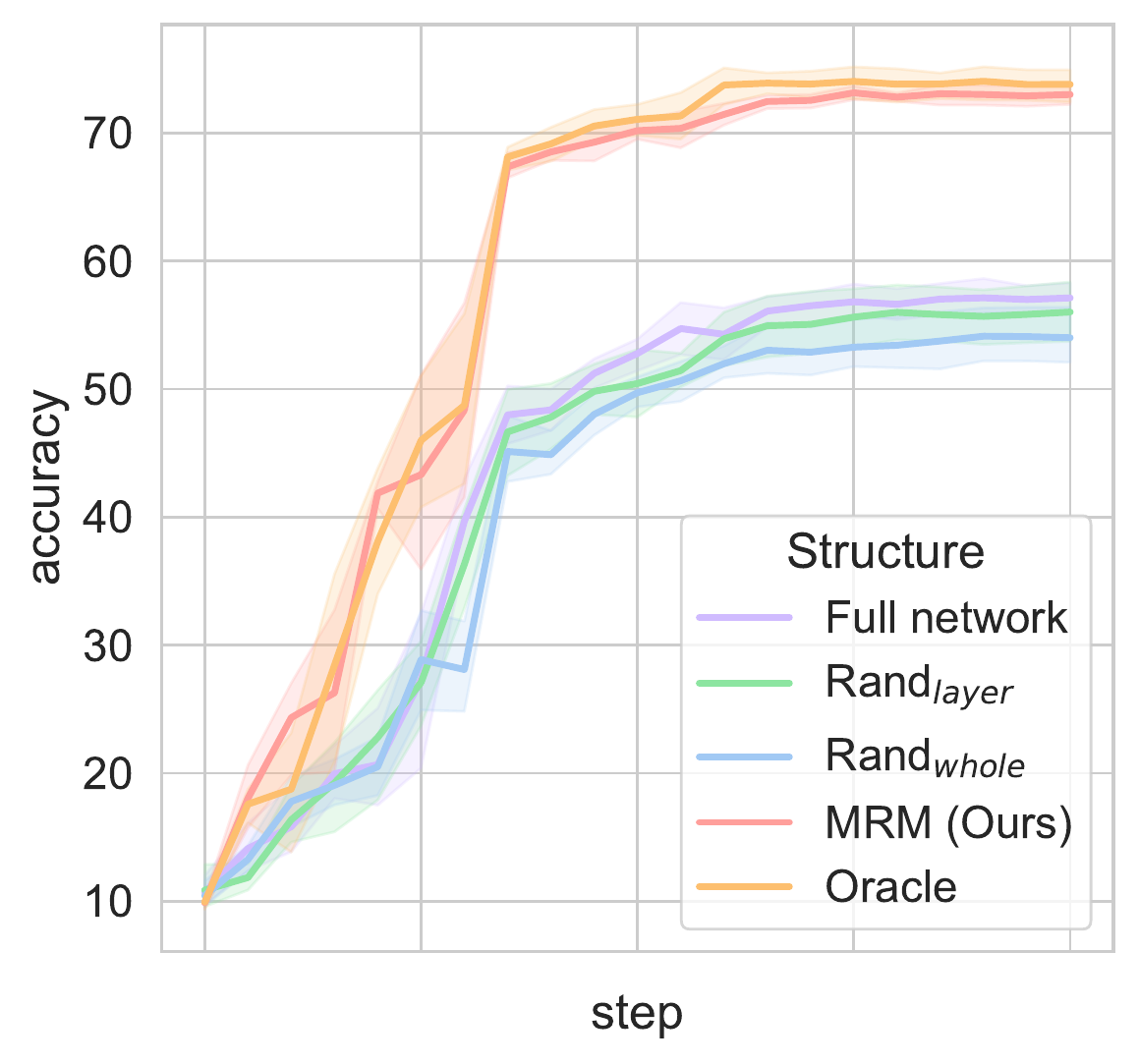}
    \hspace{0.35cm}
    \caption{OOD performance of models with different structures when trained with ERM algorithm on \textsc{FullColoredMNIST}. The oracle subnetwork and our MRM method significantly surpass the performance of full network. See details in Section~\ref{sec:lth}. }
    \label{fig:erm_lth}
\end{figure}

What causes these failures? Recent works \cite{peters2016causal, Arjovsky2019InvariantRM} argue the principle of empirical risk minimization (ERM) is at fault: if the data is generated from a fully observed causal bayesian network (CBN), then ERM would typically use all the features in the Markov blanket including those which are not the causes of the label.  It may consequently fail to perform well under distribution shifts. This is known as the out-of-distribution (OOD) generalization problem. 
In an effort to alleviate the problem, \citet{peters2016causal} proposes to first identify the target's causal parents, and constrain the model structure by only updating the parameters for the parents. 
Nevertheless, their approach is only applicable for linear problem.
\citet{sagawa2020investigation} also analyze the problem from a model structure perspective, but focusing on deep neural networks and showing that overparameterization will hurt OOD performance through data memorization and overfitting. Rather than focusing model structure, most recent works \cite{Arjovsky2019InvariantRM, sagawa2019distributionally, ahuja2020invariant, Krueger2020OutofDistributionGV, jin2020domain, koyama2020out, creager2020exchanging} mainly target improvements in the objective function over ERM.

In this work, we set out to study the effect of the model structure in OOD generalization beyond simple considerations of model capacity. 
We begin by demonstrating that even already trained models that exploit spurious correlation can contain subnetworks that capture invariant features. We then turn to investigate whether the choice of structure matters in the training process. To this end, we propose a functional lottery ticket hypothesis -- a full network contains a subnetwork that can possibly achieve \textit{better} performance for OOD generalization than full network. 
We confirm this hypothesis by experiments on a manually crafted dataset (Figure~\ref{fig:erm_lth}) with our ``oracle'' subnetwork that uses information from OOD examples. As a practical method to that avoids the use of OOD information, we propose the Modular Risk Minimization (MRM) approach. MRM is a simple algorithm to address OOD tasks via structure learning. Our approach hunts for subnetworks with a better OOD inductive bias and can also combine with other OOD algorithms, bringing consistent performance improvement.
We summarize our contributions as follows: 

\begin{itemize}
    \vspace{-0.2cm}
    \item We show that large trained networks that exploit spurious correlations contain subnetworks that are less susceptible to these spurious shortcuts.
    \item We propose a novel functional lottery ticket hypothesis: there exists a subnetwork that can achieve better OOD and commensurate in-distribution accuracy in a comparable number of iterations when trained in isolation.
    \item We propose Modular Risk Minimization (MRM), a straightforward and effective algorithm to improve OOD generalization.
    MRM helps select subnetworks and can be used in conjunction with other methods (\textit{e.g.}, IRM) and boosts their performance as well.
\end{itemize}

\section{Invariant Prediction}
\subsection{Out-of-distribution (OOD) generalization problem}
\label{sec:intro_OOD}
Consider a supervised learning setting where the data is gathered from different environments and each environment represents a different probability distribution. Let  $\left(X^{e}, Y^{e}\right) \sim \mathbb{P}^e$, where $X^e\in \mathcal{X} ,Y^e\in \mathcal{Y} $ stands for the feature random variable and the corresponding label,  $e \in \mathcal{E} = \{1, ..., E\}$ is the index for environments, and the set $\mathcal{E}$ corresponds to all possible environments. The 
set $\mathcal{E}$ is divided into two sets: seen environments  $\mathcal{E}_{\mathsf{seen}}$ and unseen 
ones $\mathcal{E}_{\mathsf{unseen}}$ ($\mathcal{E} = \mathcal{E}_{\mathsf{seen}} \cup \mathcal{E}_{\mathsf{unseen}}$). The training dataset comprises samples from 
$\mathcal{E}_{\mathsf{seen}}$. The dataset from environment $e$ is given as $D_e = \{x_i^e,y_i^e\}_{e=1}^{n^e}$, where each point $(x_i^e,y_i^e)$ is an independently identically distributed (IID) sample from $\mathbb{P}^e$ and $n^e$ is the number of samples in environment $e$. We write the training dataset as $D_{\mathsf{train}} = \cup_{e\in \mathcal{E}_{\mathsf{seen}}} D_e$.
In the rest of the work, we interchangeably use the term \textit{domain} and \textit{environment}, and we will use \textit{in}-distribution or \textit{in}-domain to refer to seen environmental data, and \textit{out}-distribution or \textit{out}-domain for unseen environmental data.

 Let $f_\theta:\mathcal{X} \rightarrow \mathcal{Y}$ denote the parametrized model with parameters $\theta \in  \Theta$. Define the risk achieved by the model as $\mathcal{R}^e(\theta)= \mathbb{E}^{e}\big[\ell(X^{e}, Y^{e})\big]$ where $\ell$ is the loss per sample (\textit{e.g}., cross-entropy, square loss). The goal of out-of-distribution (OOD) generalization problem is to learn a model that solves 
 \begin{equation}
     \min_{\theta\in \Theta}\max_{e\in \mathcal{E}}  \mathcal{R}^e(\theta).
 \end{equation}
 Since we only have access to data from $D_{\mathsf{train}}$ and do not see samples from the unseen environments, the above problem can be challenging to solve.  

\textbf{Data generation process.} 
We assume $X^e$ is generated from latent variables $Z^e = (Z_{\mathsf{inv}}^e, Z_{\mathsf{sp}}^e)$. Consider an illustrative example where $X^e$ could be the pixels in images,
while $Z_{\mathsf{inv}}$ denotes invariant features (\textit{e.g.}, foreground) and  $Z_{\mathsf{sp}}$ denotes spurious features (\textit{e.g.}, background).
 We write $X^e = G(Z_{\mathsf{inv}}^e, Z_{\mathsf{sp}}^e)$, where $G$ is a map from the latent space to the pixel space.  $Y^e$ is the label for the object and it is determined based on the following map $Y_e = F(Z_{\mathsf{inv}}^e)$. The combination pattern of $Z_{\mathsf{inv}}^e$ and $Z_{\mathsf{sp}}^e$ varies across domains, hence generating different environmental distributions. In our description of the data generation, we do not use noise variables to keep things simple ($Y$ is related to $Z$ deterministically and $X$ is related to $G$ deterministically). 
Suppose that we can recover $Z_{\mathsf{inv}}^e$ and $Z_{\mathsf{sp}}^e$  from $X^e$ and we write these inverse maps as $Z_{\mathsf{inv}}^e = G^{\dagger}_{\mathsf{inv}}(X^e)$ and $Z_{\mathsf{sp}}^e = G^{\dagger}_{\mathsf{sp}}(X^e)$. The ideal function that the model wants to learn is $F \circ G_{\mathsf{inv}}^{\dagger}$  as it yields zero error and only relies on invariant latents. 
However, as we explain next that due to selection biases the model can often find it hard to learn a model that only relies on $Z_{\mathsf{inv}}^e$. 

\textbf{Bias.} To explain why the datasets have a bias, let us consider a simple example, where $Z_{\mathsf{inv}}^e \in \{-1,1\}^{D_{\mathsf{inv}}}$, $Z_{\mathsf{sp}}^e \in \{-1,1\}^{D_{\mathsf{sp}}}$ and $Y^e\in \{-1, 1\}$.  Suppose  each component of $Z_{\mathsf{inv}}^e$ is $Y^e$  and each component of $Z_{\mathsf{sp}}^e$ independently takes a value equal to $Y^e$ with a probability $p^e$ and $-Y^e$ with a probability $1-p^e$. If $p^{e}$ is close to $1$ and $G_{\mathsf{sp}}^{\dagger}$ is an easier function to learn than $G_{\mathsf{inv}}^{\dagger}$, then it is intuitive that the model can instead learn $Z_{\mathsf{sp}}^e$ and predict the label $Y_e$. However, this can be catastrophic as the correlation between the spurious feature and the label only holds in the training environments and does not translate to the test environments where $p_e=\frac{1}{2}$. 
Even if $p^e$ is small, as long as $Z_{\mathsf{sp}}^e$ is high dimensional ($D_{\mathsf{sp}} \gg D_{\mathsf{inv}}$), the model can be shown to significantly rely on $Z_{\mathsf{sp}}^e$ \cite{nagarajan2020understanding}. The above example uses binary valued latents for ease of exposition, but the same biases can occur in more general settings where the same problems plague the models. 


\subsection{A Motivating Example}
In this section, we use a simple example to motivate the constraints we impose in our approach.
Consider the data setting described in the previous section, $Z_{\mathsf{inv}}^e \in \{-1,1\}$ ($D_{\mathsf{inv}}=1$) and $Z_{\mathsf{sp}}^e \in \{-1,1\}^{D}$ ($D_{\mathsf{sp}}=D$).  We take $G$ to be the identity map as in \citet{tsipras2019robustness, rosenfeld2020risks} and thus $X^e = (Z_{\mathsf{inv}}^e,Z_{\mathsf{sp}}^e)$.  Suppose the model $f_\theta$ is a linear predictor; we refer to the components associated with invariant feature as $w_{\mathsf{inv}}$ and those associated with the spurious feature as $\boldsymbol{w}_{\mathsf{sp}}$. 

\textit{Learning a sparse classifier:} Find a maximum margin classifier that satisfies the following sparsity constraint: the number of non-zero coefficients $\leq d$. We denote such a classifier as $f_{\mathsf{sparse}}^{d}$. 

In the next proposition, we compare the behavior of the sparse classifier that we defined above with a classifier that relies only on spurious features. We construct a regular 
classifier $f_{\mathsf{reg}}$ (with unit norm) that purely relies on the spurious features, i.e., $w_{\mathsf{inv}}=0$ and  $\boldsymbol{w}_{\mathsf{sp}} = \boldsymbol{1}\frac{1}{\sqrt{D_{\mathsf{sp}}}}$ and thus has poor OOD performance.  
We denote the average error rate of the classifier $h$ on seen (or unseen) environments as $\mathsf{Err}_{\mathsf{seen}}(h)$ (or $\mathsf{Err}_{\mathsf{unseen}}(h)$.) Here the error for binary classification is defined to be $\mathsf{Err}^e(h) = \frac{1}{2}\E_{\left(X^{e}, Y^{e}\right) \sim \mathbb{P}^e}\left[1 - Y^e h(X^e)\right]$. We denote the margin of classifier for data in environment $e$ as $\mathsf{Margin}^e$.

\begin{proposition}
\label{prop:linear}
Consider the dataset in Section~\ref{sec:intro_OOD} with $Z_{\mathsf{inv}}^e \in \{-1,1\}$ ($D_{\mathsf{inv}}=1$) and $Z_{\mathsf{sp}}^e \in \{-1,1\}^{D}$ ($D=D_{\mathsf{sp}}$). 
Let $n$ be the number of training samples in $D_{\mathsf{train}}$, $c$ be a constant in $(0, 1)$ such that for all $e\in \mathcal{E}_{\mathsf{seen}}$, $p^e > \frac{1}{2} + \frac{c}{2}$ and $p^e=\frac{1}{2}$ for $e\in \mathcal{E}_{\mathsf{unseen}}$. For sparsity constraint $d=2$, we have:

\vspace{-0.5cm}
\begin{itemize}
\item  Compare margin for in-distribution sample: for any $\delta \in (0,1)$, if $D\geq \frac{1}{2c}\sqrt{2ln(n) / \delta}$, then with a probability at least $1-\delta$, $\mathsf{Margin}_{\mathsf{seen}}^e(f_{\mathsf{sparse}}^{d}) < \mathsf{Margin}_{\mathsf{seen}}^e(f_{\mathsf{reg}}^{}) $;
\item Similar in-distribution performance $\forall e \in \mathcal{E}_{\mathsf{seen}},$ $\mathsf{Err}_{\mathsf{seen}}^e(f_{\mathsf{sparse}}^{d}) = 0$, $\mathsf{Err}_{\mathsf{seen}}^e(f_{\mathsf{reg}})\leq 2 e^{-2c^2D}$;
\item Better out-distribution performance: $\forall e \in \mathcal{E}_{\mathsf{unseen}}$, $\mathsf{Err}_{\mathsf{unseen}}^e(f_{\mathsf{sparse}}^{d}) = 0$ and $\mathsf{Err}_{\mathsf{unseen}}^e(f_{\mathsf{reg}})=0.5$.
\end{itemize}
\end{proposition}

From the above Proposition, we can conclude that if  $c$ or $D$ is high, then the train accuracy of the sparse classifier and the regular classifier are similar but the OOD accuracy of the two classifiers are different with sparse classifier being much better. 
The algorithm is likely to select the regular classifier over the sparse classifier as it has a much higher margin than the sparse classifier. 


Proposition 1 compares the optimal sparse classifier with a purely spurious one. Both have same in-distribution performance, but the former has a better OOD performance.
We compare the margins to show that if we use a gradient descent on logistic loss, it will be biased towards the spurious classifier \cite{soudry2018implicit}.
We clarify that Proposition 1 is not intended to show a tradeoff between OOD performance and margin. 
Consider the experiment of spiral vs. linear boundary of Sec 3.1 in \citet{parascandolo2020learning}.
In the experiment, the spiral boundary is associated with invariant features and the linear boundary is associated with spurious ones. The authors set the margin for linear boundary to be larger than the that of the spiral boundary. In this case, ERM learns a model that uses spurious features. Even if we were to reduce the margin of the linear boundary to be smaller than the spiral boundary, ERM continues to rely on the spurious features as it prefers to use a simpler margin \cite{shah2020pitfalls}.



For this linear setting that we discussed above,
we can learn a constrained max-margin classifier by adding $\ell_1$ constraints. This is a tractable problem to solve as the problem remains convex. However, as we move to neural networks, learning sparse classifiers with good OOD performance  is significantly more challenging owing to the non-convexity. This issue is the subject of later sections.
Before we address this issue of learning sparse networks with good OOD properties, there is another important question to be answered. In the setting of the above proposition, we rely on the fact that a sparse model exists that relies on invariant features only and yields better OOD performance. How do we do know that this is a proper assumption for real datasets used for neural network training?  In the next section, we analyze neural networks via modular subnetwork introspection to show that such a sparse model exists.

\section{A Functional Modularity Based Analysis}
\subsection{Preliminaries}
\label{sec:preliminaries}
\textbf{Technical approach.\ } 
The modularity property of neural network has long been considered as an essential foundation of systematical generalization \cite{ballard1987modular, marcus1998rethinking, csordas2020neural}. 
Consider a task that can be compositionally separated into different independent subtasks, we aim to probe a functional module subpart of the full neural network that can solve one particular subtask.
Following \citet{zhou2019deconstructing, csordas2020neural}, we identify different subnetworks which perform different functions, from a given pretrained network.


 Specifically, we deem functional modules to be particular subsets of the weights inside a neural network.
For a $L$ layer neural network model $f(\mathbf{w}_1, \cdots, \mathbf{w}_L; \cdot)$ where $\mathbf{\theta} = \{\mathbf{w}_1, \cdots, \mathbf{w}_L\}$, we model the subnetwork with a set of binary masks $\mathbf{m}_l \in \{0, 1\}^{n_l}$ on the $l$-th layer weight tensor $\mathbf{w}_l \in {\R}^{n_l}$, where ${n_l}$ is the number of dimensionality of the $l$-th layer network parameters. The subnetwork is then given by $f(\mathbf{m}_1\odot \mathbf{w}_1, \cdots, \mathbf{m}_L\odot \mathbf{w}_L; \cdot)$. Further, in order to make this subnetwork structure learnable, we assume each entry of the mask to be independent Bernoulli random variables, and model their logits as $\boldsymbol{\pi}_l \in {\R}^{n_l}$. Hence, in this probabilistic modeling setting, the $l$-th layer subnetwork structure $\mathbf{m}_l$ is generated by performing Bernoulli sampling with parameters $\mathrm{sigmoid}(\boldsymbol{\pi}_l)$. We adopt Gumbel-sigmoid trick \cite{jang2016categorical} to enable an end-to-end training process, together with a logit regularization term to promote subnetwork sparsity \cite{csordas2020neural}. For each particular subtask, our analysis will output a 
logits tensor for each neuron in the form of $\boldsymbol{\pi} = \{\boldsymbol{\pi}_1, \cdots, \boldsymbol{\pi}_L\}$, and thereby uncover the corresponding functional module within the neural network in the form of binary tensor $\mathbf{m} = \{\mathbf{m}_1, \cdots, \mathbf{m}_L\} = \{\text{sigmoid}(\boldsymbol{\pi}_l) > 0.5 \ | \ l = 1,2,\cdots\}$.
We then use the term \textit{modularity probing} method to refer to this technique subsequently.
We will interchangeably use the term of \textit{module} and \textit{subnetwork} due to their consistency in our context. 

\begin{figure}[t]
    \subfigure[Accuracy of baselines.]{
    \begin{minipage}{3.5cm}
    \includegraphics[width=4cm]{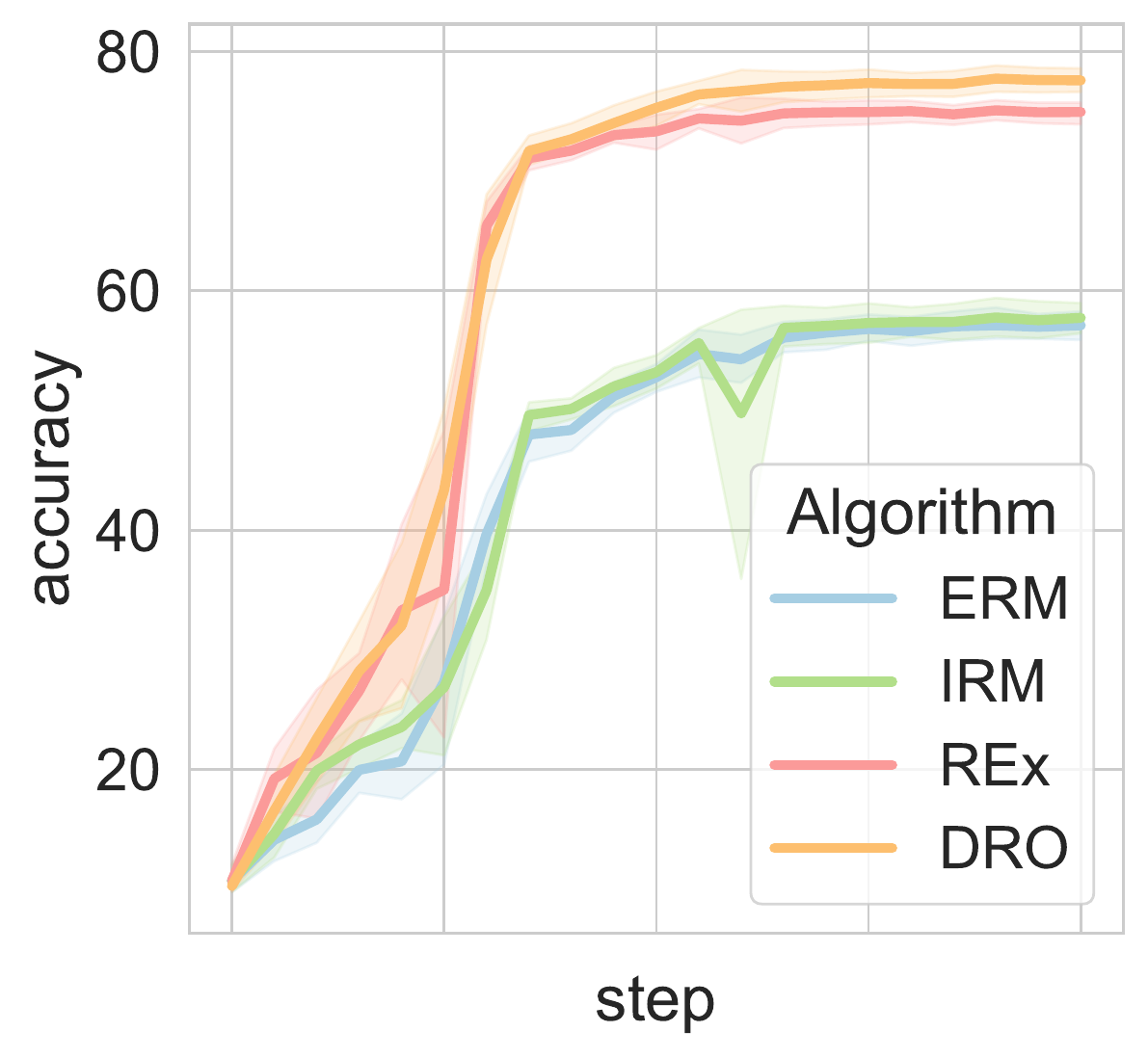}
    \end{minipage}
    \label{fig:baselines_acc}
    }
    \hspace{0.1cm}
    \subfigure[Accuracy of module in ERM.]{
    \begin{minipage}{4cm}
    \includegraphics[width=4cm]{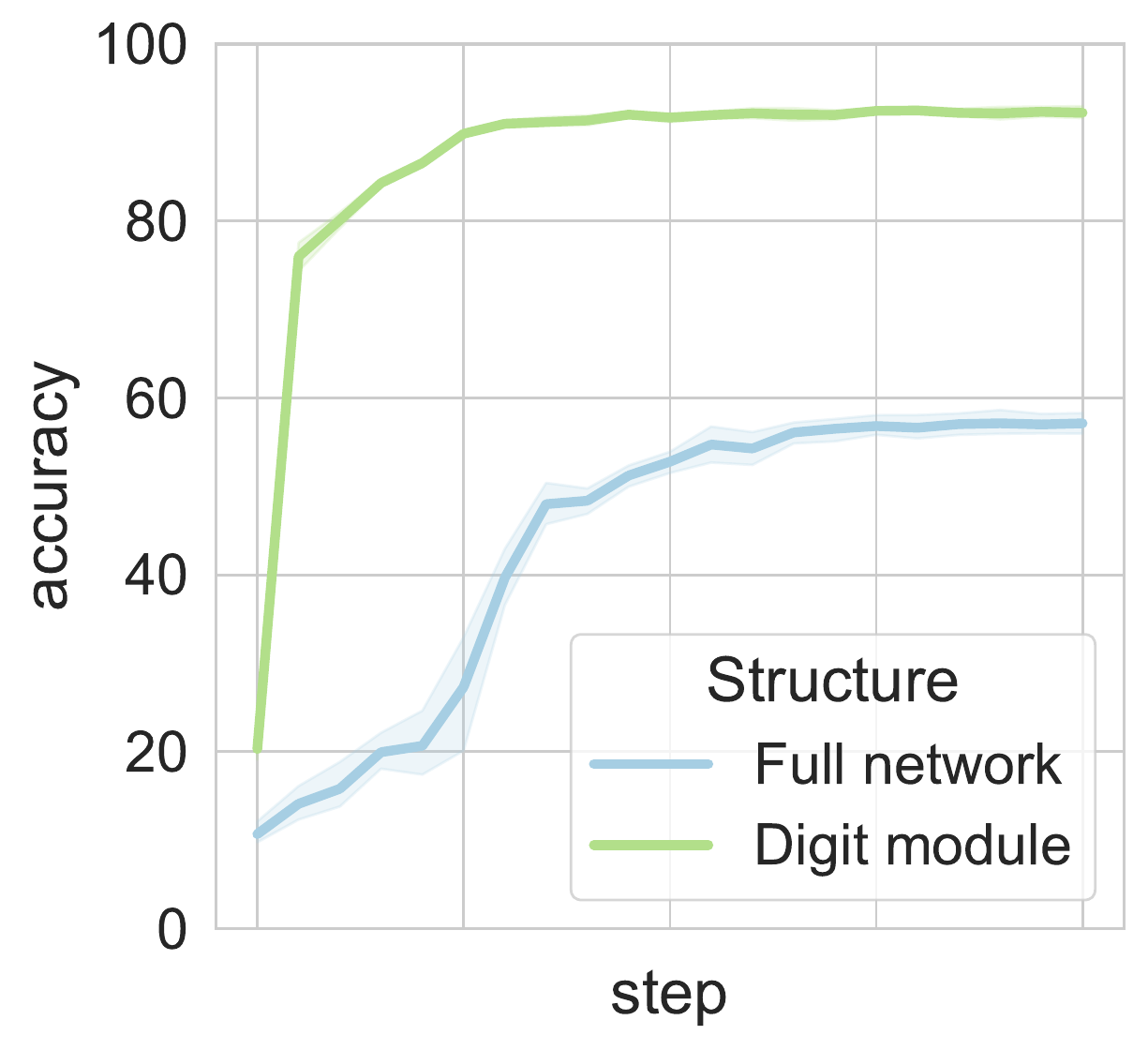}
    \end{minipage}
    \label{fig:digit_acc}
    }
    \caption{\textit{Left}: OOD accuracy for four algorithms. \textit{Right}: OOD accuracy for ERM algorithm and its digit module. The plot shows that a highly biased model can contain an unbiased subnetwork.}
    \label{fig:universal_digit_acc}
    \vspace{-0.5cm}
\end{figure}

\textbf{Dataset construction.\ }
We take the intuition from \citet{Arjovsky2019InvariantRM, nam2020learning,ahuja2020empirical, ahmed2021systematic} to design a biased variant of the original MNIST dataset \cite{lecun1998gradient}. A discussion about the difference between ours and theirs is deferred to supplementary materials. 
The digit shape semantics are considered as $Z_{\mathsf{inv}}$ while color semantics as $Z_{\mathsf{sp}}$. We choose ten different kinds of color and define a one-to-one corresponding bias relationship with ten digit class (\textit{e.g.}, ``2" $\leftrightarrow$ ``green", ``4" $\leftrightarrow$ ``yellow"
). 
For each domain, we define the bias coefficient to be the ratio of the data that obeys this relationship. Those images which don't follow this relationship are then assigned with random colors. The bias coefficient for two in-domains is $(1.0, 0.9)$ respectively, 
which means the first domain is completely biased and 90 percent of the second domain is biased. 
For the out-domain, all images are assigned a random color for evaluating to how much extent the model has learned the invariant feature. The out-domain will serve as a tool environment for module learning in this section, representing a thorough disentanglement of two attributions. It will then act as the test distribution in a realistic setting in Section~\ref{sec: practical}.
Unless otherwise specified, the label is set as the class where the invariant attribution lies.
We use the term \textsc{FullColoredMNIST} to refer to this 
task to distinguish with the binary colored mnist dataset in \citet{Arjovsky2019InvariantRM}.

\textbf{Algorithms analyzed.}
We study four OOD generalization algorithms in this paper: Empirical Risk Minimization (ERM) \cite{vapnik1999overview}, Invariant Risk Minimization (IRM) \cite{Arjovsky2019InvariantRM}, Risk Extrapolation (REx) \cite{Krueger2020OutofDistributionGV} and group Distributional Robust Optimization (DRO) \cite{sagawa2019distributionally}. More details about them are left to supplementary materials.
Figure~\ref{fig:baselines_acc} plots the generalization performance of these algorithms w.r.t. the training process. REx (76.17\%) and DRO (78.56\%) methods surpass ERM baseline by a large margin, while IRM (59.55\%) only gets slightly better results than ERM (58.04 \%).
The failure of IRM in realistic problems has been analyzed in \citet{jin2020domain, nagarajan2020understanding, rosenfeld2020risks, ahuja2020empirical} and attributed to the overparameterization regime and curse of dimensionality, hence we omit related discussion here.

\subsection{Modular subnetwork introspection}
\label{sec:instropection}

Departing from previous
approaches, in this section we think of learning the digit and color semantics as different functional subtasks of the original task, rather than opposite non-spurious / spurious features. We split the out-domain into two parts and refer to them as the \textit{in}-split and \textit{out}-split of the \textit{out}-domain (terminology from \citet{gulrajani2020search}). 
We define two subtasks, identification of digit and identification of color. For each subtask, we assume that we have access to respective semantic labels. It's important to note that the semantic color label is used here for analysis and is not a part of our main method described later.

 
In order to study the functional module for the two subtasks, we apply the modularity probing method to diagnose given pretrained models. 
Specifically, we separately train and get a digit and a color subnetwork for each model across different algorithms and training steps.
We evaluate the obtained digit modules' behaviors on the out-split of out-domain (as the in-split has been taken for module searching). 
Figure~\ref{fig:digit_acc} suggests a significant evidence that, even for biased models such as ERM trained ones, there exist unbiased invariant subnetworks (digit modules) with good OOD generalization ability. 
We also explore this property for other modules and algorithms and defer these results to supplementary materials.

\begin{figure}[t]
    \centering
    \hspace{0.4cm}
    \includegraphics[width=0.6\columnwidth]{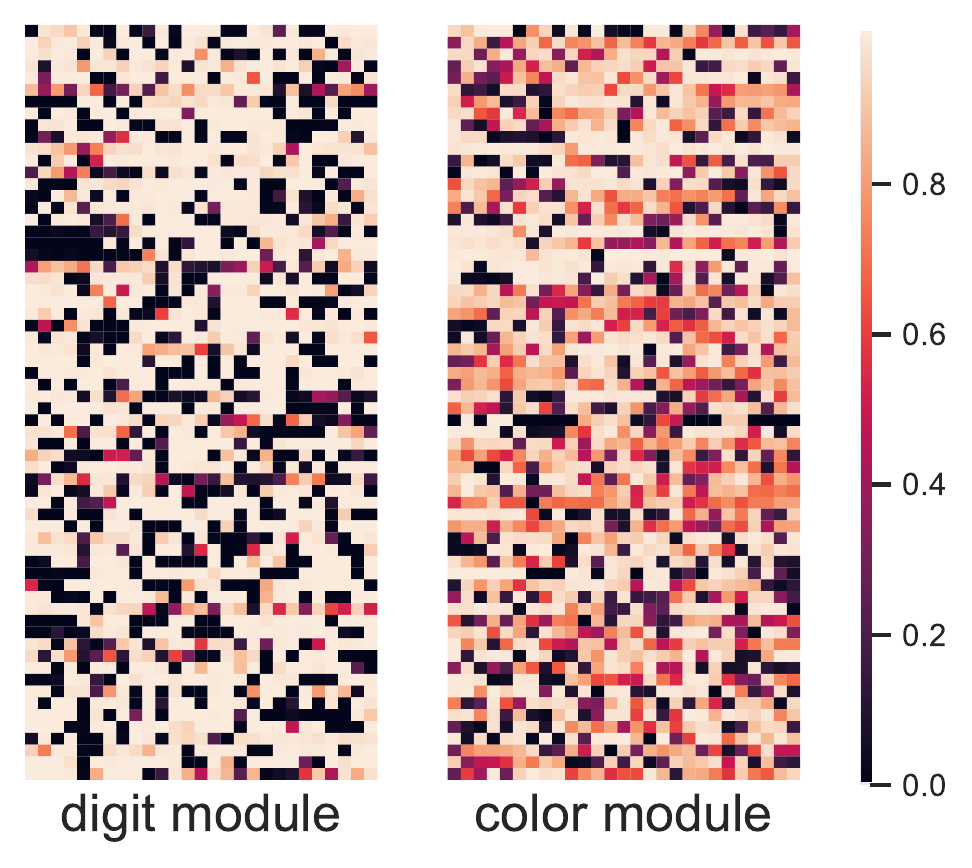}
    \caption{The visualization of the Bernoulli probability of digit and color functional module for the first (convolutional) layer. The weight tensor is reshaped to two dimension for display convenience. The probability takes value from 0 to 1.}
    \label{fig:visualize}
    \vspace{-0.5cm}
\end{figure}

\textbf{Discussion about the sparsity of digit weights / features.}
We additionally visualize the Bernoulli probability of learned subtask modules.
Figure~\ref{fig:visualize} displays the first layer of model trained on two in-domains with ERM. We can see that the color feature is more pervasive than digit feature, spreading over a broader range across the neurons. 
Although the sparsity of weights is not exactly the sparsity of features, the discovery 
is aligned with
the assumption in Proposition~\ref{prop:linear} that $D_{\mathsf{inv}}$ has a small number of dimensionality.
The visualization results of other layers are similar to this, and can be found in the supplementary materials.


In this section we confirm in large \textit{trained} models, there lie invariant functional modules, \viz\ subnetwork that behaves well for target invariant function (\textit{e.g.}, digit classification). 
However, it's more worthwhile to find out about whether an appropriate subnetwork structure can help improve in the sense of OOD generalization \textit{during training}. We investigate this problem in the next section.

\section{Structure Matters: Towards Functional Lottery Tickets Hypothesis}
\label{sec:lth}

\citet{frankle2018lottery} proposes the lottery ticket hypothesis from a pruning perspective, suggesting that among all different subnetworks, there exists a so-called ``winning ticket" that can \textit{reach} the generalization ability of the full network with faster training speed. In this original lottery ticket hypothesis, the data distribution remains unchanged across training and testing. Whereas in our OOD context, we seek a model subnetwork whose functional predictions are invariant with respect to the change in distribution.


\textbf{The functional lottery ticket hypothesis:} {A randomly initialized, dense neural network contains a subnetwork that is initialized such that — when trained in isolation — it can achieve \textit{better} out-of-distribution performance 
w.r.t. the given function (\textit{e.g.}, digit identification in our context)
than the original full network after training for the same number of iterations.
}

Concretely, our functional lottery ticket hypothesis claims that for a dense neural network model $f(\mathbf{w};\mathbf{x})$ with initialization parameter $\mathbf{w}_0$, there exists a module $\mathbf{m}$ enabling a subnetwork $f(\mathbf{m}\odot\mathbf{w};\mathbf{x})$ to \textit{surpass} the OOD performance of full network when trained from $f(\mathbf{m}\odot\mathbf{w}_0;\mathbf{x})$ on in-distribution data. 
Note that this is a stronger statement than the original lottery ticket hypothesis, which only requires the winning ticket to reach similar performance to the full network in the IID setting.

\textbf{Demonstration.} 
How do we identify the functional winning tickets? How should one search for a structure that is best for OOD generalization?
To unravel the possible best result one can reach, we design an ``oracle" subnetwork. After obtaining an ERM trained model, the structure of oracle module is found with the aid of the in-split data part of out-domain. Namely, we use the modularity probing technique introduced in Section~\ref{sec:preliminaries} with these ``oracle" data from the out-distribution, and then deploy the resulting subnetwork back onto the previous initialization. We then evaluate all methods on the out-split.

According to \citet{sagawa2020investigation}, the underparameterized regime can keep the model from overfitting to spurious features. Therefore, we choose the random subnetwork as another option to investigate whether sparsity / underparametrization alone can achieve an unbiased solution.
``\text{rand}$_{\text{whole}}$" method keeps the ratio of full network same as the oracle subnetwork. In other works it has been claimed that the sparsity per layer is the only working factor for pruning \cite{su2020sanity, frankle2020pruning}, we also experiment with the ``\text{rand}$_{\text{layer}}$" method, where we randomly sample subnetworks with the same per-layer-sparsity.


\begin{figure}[t]
    \centering
    \subfigure{
    \begin{minipage}{3.5cm}
    \includegraphics[width=4cm]{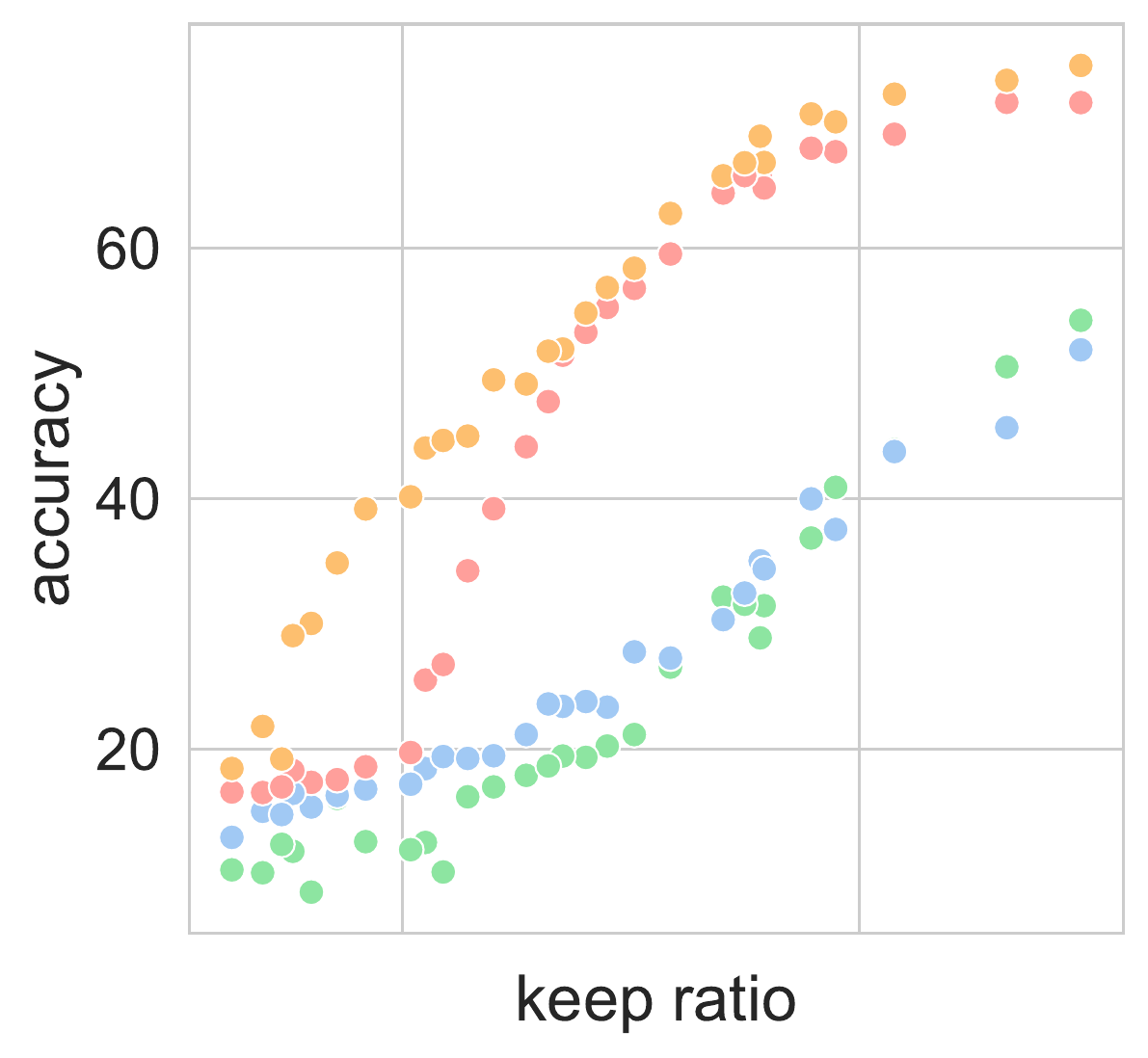}
    \end{minipage}
    \label{fig:lth_sparsity}
    }
    \hspace{0.2cm}
    \subfigure{
    \begin{minipage}{4cm}
    \includegraphics[width=4cm]{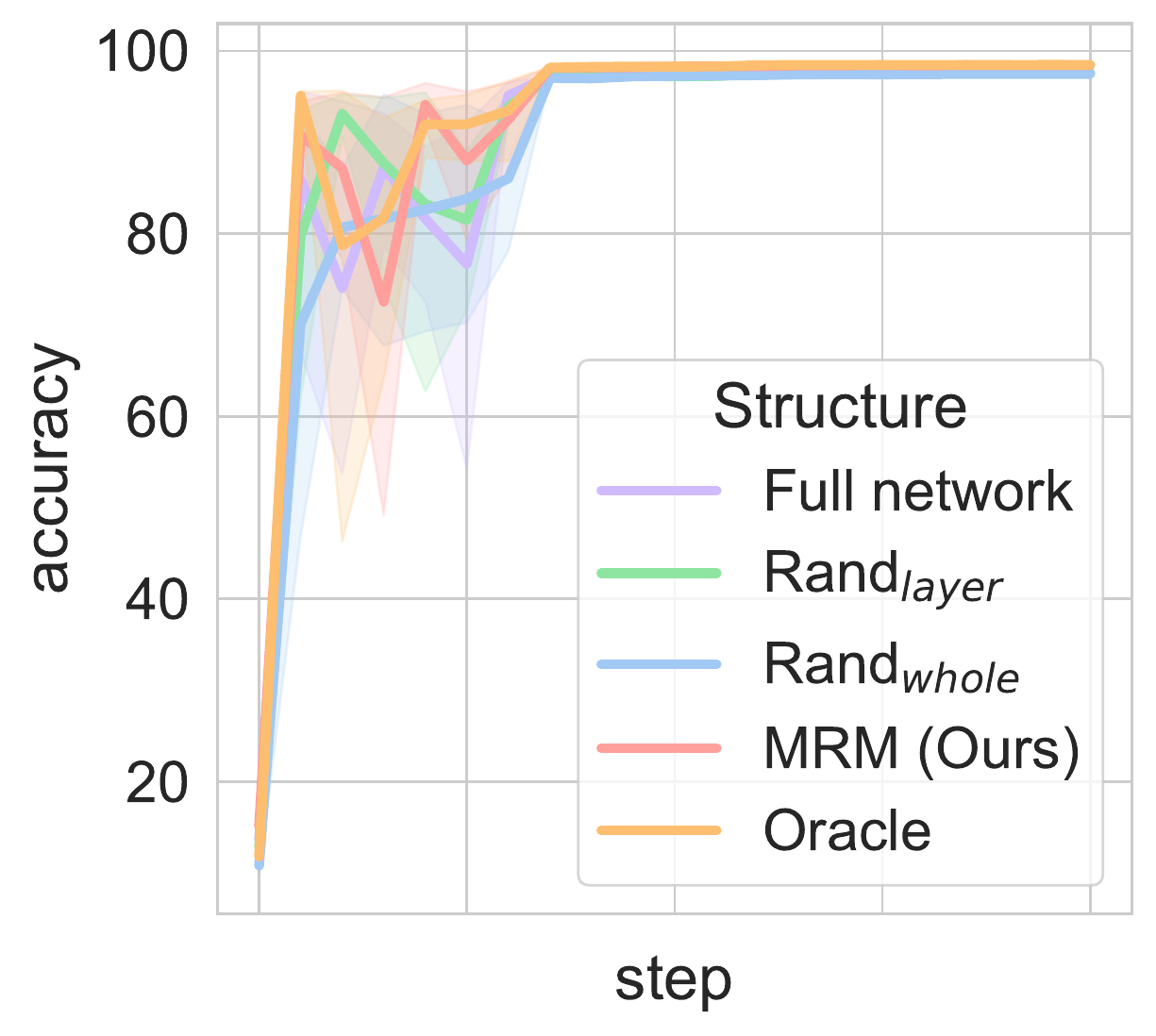}
    \end{minipage}
    \label{fig:erm_lth_indistribution}
    }
    \caption{\textit{Left:} Performance of different networks for various levels of sparsity. Here the \emph{keep ratio} is defined to be $1 -$ sparsity and left side of figure means smaller keep ratio. \textit{Right:} In-distribution generalization performance of different subnetworks in Section~\ref{sec:lth}.}
    \vspace{-0.4cm}
\end{figure}

Figure~\ref{fig:erm_lth} and \ref{fig:erm_lth_indistribution} show the OOD and in-distribution generalization results for these subnetworks with ERM training on the \textsc{FullColoredMNIST} dataset respectively.
The oracle subnetwork beats the original ERM by a large margin for OOD and maintains indistinguishable performance for the in-distribution examples, confirming that a good module structure can indeed surpass the full network in terms of this digit function. 
We show their performance with greater levels of sparsity in Figure~\ref{fig:lth_sparsity} and see a considerable consistent accuracy gap between oracle and random baselines for all level of sparsity.
The validity of our functional lottery ticket hypothesis is thereby empirically affirmed, and we thus propose that \textit{appropriate structure induction can impose a needed inductive bias to prevent the model from fitting the spurious correlation}.
We also conclude that sparsity constraint imposed cannot help alone, as two random methods don't yield non-trivial benefit than ERM (both under 60\% accuracy). Additionally, we demonstrate our hypothesis is also applicable for other OOD algorithms in the supplementary materials.

\textbf{Discussion.}
One can rightly criticize this investigation as unfair in that it compares a method using out-domain data to baselines without such privileged access. We acknowledge this issue and, for now, only seek out this ``oracle subnetwork" to highlight the importance of structure. 
We now turn to the question of how we can design a practical structure searching algorithm to overcome this limitation.

\subsection{Modular risk minimization}
\label{sec:mrm_methodology}

The motivation behind the Modular Risk Minimization (MRM) method is to get rid of spurious features by hunting for a desired functional winning ticket. 
Since we have shown in previous subsection that 
contrary to \citet{sagawa2020investigation},
only sparsity constraints imposed at the beginning of training cannot do the magic,
we propose the following criterion:

\textit{A good structure should balance the predictiveness for invariant feature and sparsity well.}

Our procedure first trains the model with ERM resulting in a potentially biased classifier.
At this time, the functional lottery ticket hypothesis suggests that the model has already learned a promising functional module within.
Hence, we simply apply the subnetwork probing technique with \textit{training} data to learn the potential advantageous structure.
The structure learning objective takes a combination of cross entropy loss and sparsity regularization to balance the two desiderata mentioned above.
We then simply train from scratch again only with the weights in the obtained subnetwork and fix the other weights to zero. We summarize this procedure in Algorithm~\ref{alg:mrm}, where $i, c, l$ are respectively the index for a datum, label class and network layer.
It's notable that in Figure~\ref{fig:erm_lth}, our proposed MRM algorithm successfully reaches a very close accuracy to the oracle optimal structure.

\begin{algorithm}[t]
   \caption{Modular Risk Minimization}
   \label{alg:mrm}
\begin{algorithmic}
   \STATE {{\bfseries Input}: Data $\{(x^e_i, y^e_i)\}_{i, e}$, neural network $f(\mathbf{w}; \cdot)$, subnetwork logits $\boldsymbol{\pi}$, the coefficient of sparsity penalty $\alpha$, number of steps for model and subnetwork structure training $N_1, N_2$.} 
  \\~\\
\vspace{-0.2cm}
\STATE{\textbf{Stage 1}:\ full model (pre-) train}
\STATE{Get model initialization $\mathbf{w}_0$}.
\FOR{n=1 \textbf{to} $N_1$}
\STATE{Update $f$ with $\mathcal{L}_{\text{CE}}(\mathbf{w}) := \sum_{i,c} y_{i, c}\log f(\mathbf{w}; x_i)_c $.}
\ENDFOR
\STATE{\textbf{Stage 2}:  module structure probing}\\
\FOR{n=1 \textbf{to} $N_2$}
\STATE Sample subnetwork $\mathbf{m}$ $\sim$ $\text{sigmoid}(\boldsymbol{\pi})$.\\
Update module $\boldsymbol{\pi}$ with \\
\qquad $\mathcal{L}_{\text{MOD}} = \mathcal{L}_{\text{CE}}(\mathbf{m}\odot\mathbf{w}) + \alpha \sum_{l,j} \boldsymbol{\pi}_{l, j}$.
\ENDFOR
\STATE \textbf{Stage 3}: subnetwork retrain\\
\STATE Obtain the module by hard thresholding:\\
\qquad\quad$\mathbf{m} = \{\boldsymbol{\pi}_l > 0 \ | \ l=1,2, \cdots \}$.\\
\STATE{Set model parameters back to $\mathbf{w}_0$.}
\FOR{n=1 \textbf{to} $N_1$}
\STATE Update $f$ with  $\mathcal{L}_{\text{CE}}(\mathbf{m}\odot\mathbf{w})$.\\
\ENDFOR

\end{algorithmic}
\end{algorithm}

\textbf{Structure learning by invariance capturing.}
Notably, MRM \textit{does not impose any invariance across domains} and is thus orthogonal to the advantage of other OOD algorithms. 
Unlike heuristic structure searching paradigms \cite{lee2018snip, wang2020picking}, our method can incorporate \textit{any} OOD generalization approach to help improve the structure learning and model training process.
This enables MRM to act as a plug-in method to boost other algorithms by supplying a good subnetwork learned from their respective objectives. 
We simply replace the cross entropy loss $\mathcal{L}_{\text{CE}}$ in Algorithm~\ref{alg:mrm} with recently developed OOD losses: $\mathcal{L}_{\text{IRM}}$, $\mathcal{L}_{\text{REx}}$ and $\mathcal{L}_{\text{DRO}}$ by IRM, REx and DRO. These new methods are therefore referred by Modular Invariant Risk Minimization (ModIRM), Modular Risk Extrapolation (ModREx) and Modular Distributionally Robust Optimization (ModDRO).
With these explicitly OOD learning algorithms, the cross domain variance is taken into account and thereby ameliorates the invariance property of the subnetwork. We note that more flexible combinations can be explored (\textit{e.g.}, use a different loss design for subnetwork and model learning), but we leave this for future work and only study these three variants in this paper.


\subsection{Ablation for winning tickets learning}
To better understand
the crucial succeeding reasons for two kinds of winning tickets -- oracle subnetworks and MRM subnetworks, we conduct corresponding ablation studies.

\begin{figure}[t]
    \centering
    \subfigure{
    \begin{minipage}{3.5cm}
    \includegraphics[width=4cm]{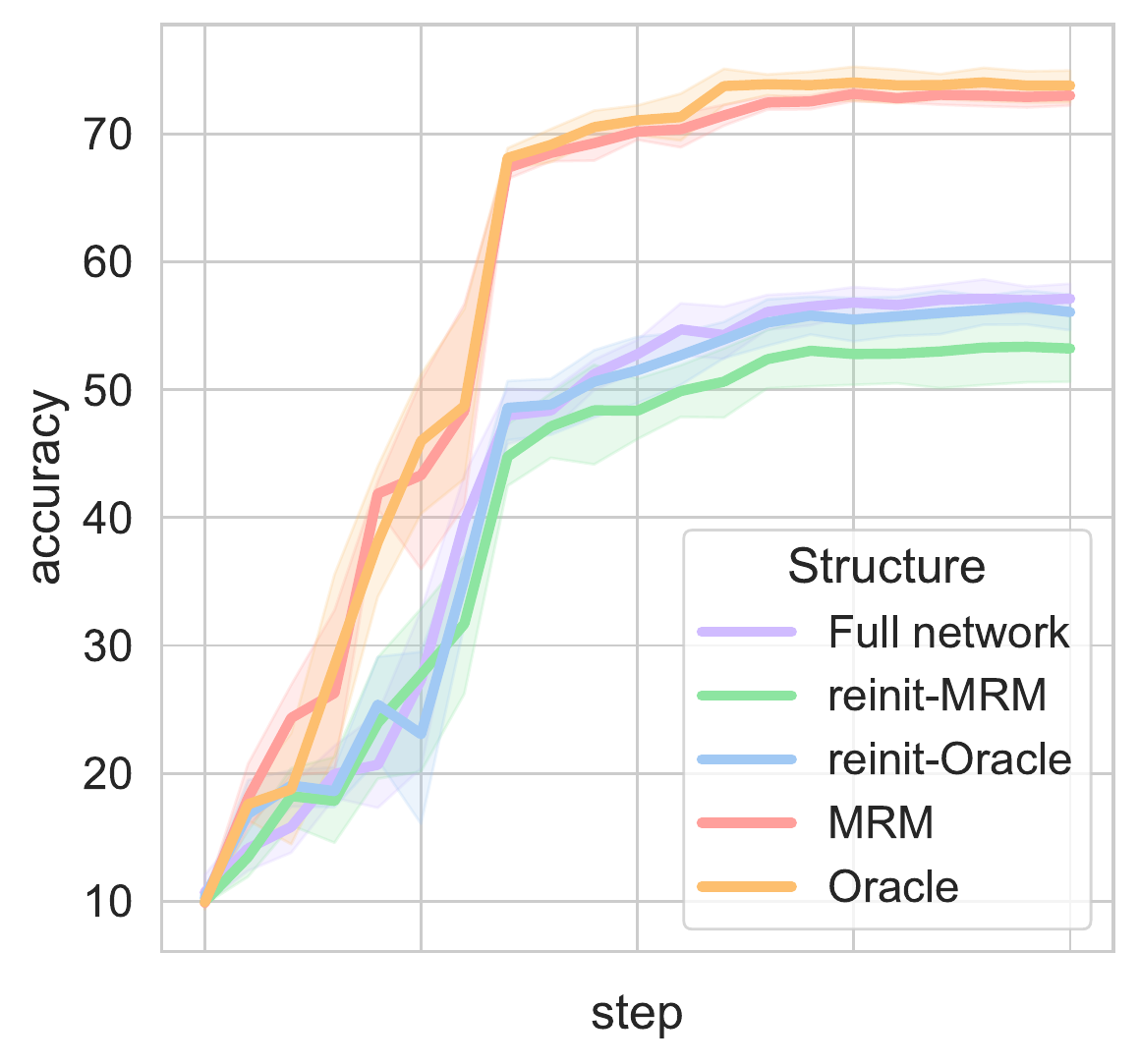}
    \end{minipage}
    \label{fig:reinit_lth}
    }
    \hspace{0.2cm}
    \subfigure{
    \begin{minipage}{4cm}
    \includegraphics[width=4cm]{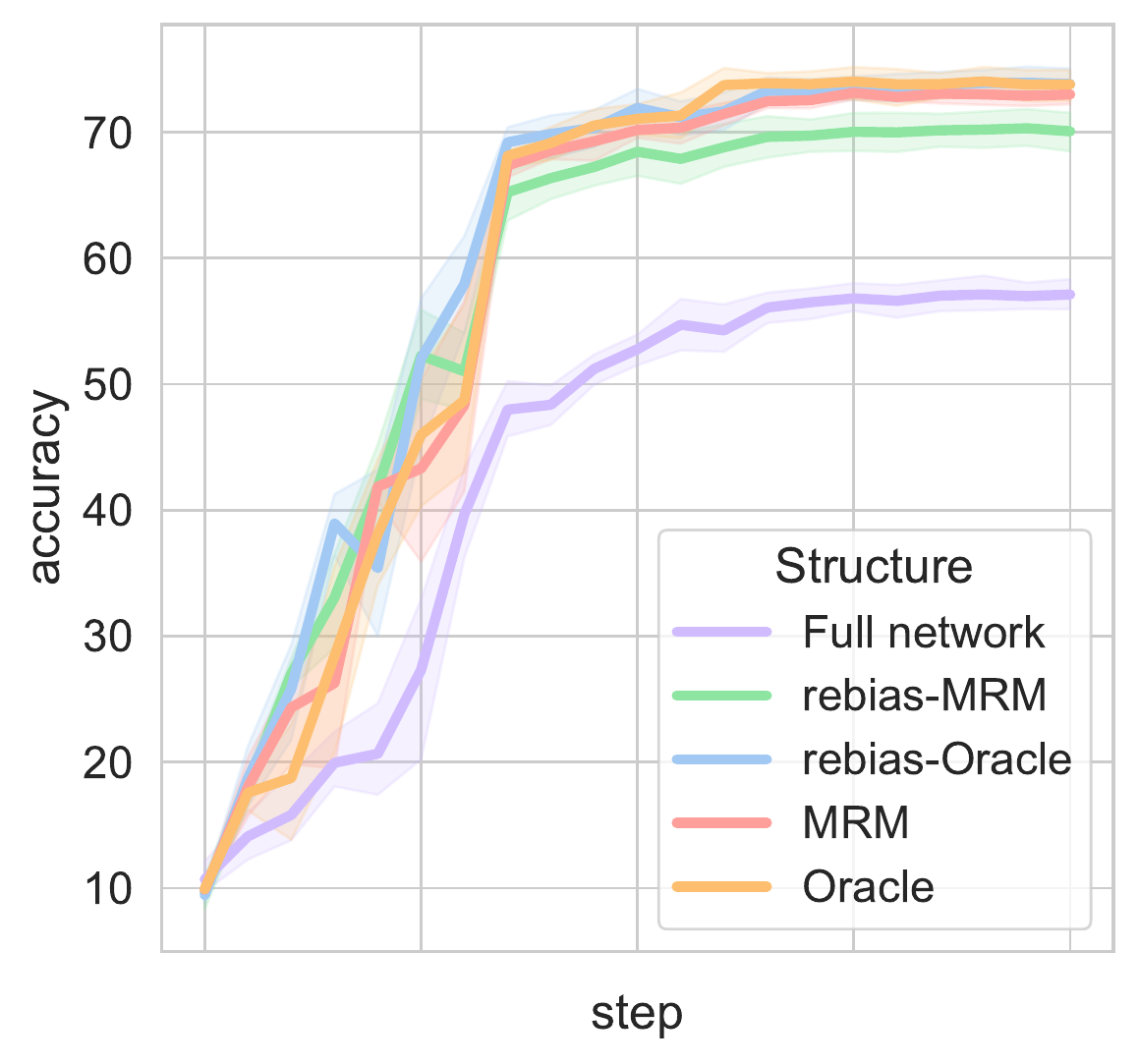}
    \end{minipage}
    \label{fig:reseed_lth}
    }
    \caption{\textit{Left:} Ablation for the importance of initialization. With re-initialized model weights, the winning tickets fail to win the jackpot. \textit{Right:} Rearrange the color-digit relationship slightly reduces performance.}
\end{figure}

\textbf{Importance of initialization.}
One of the main argument of \citet{frankle2018lottery} is 
that the winning tickets cannot be learned effectively 
without its original initialization.
We verify this for our hypothesis as well.
Figure~\ref{fig:reinit_lth} depicts the failure of functional winning tickets when random \textit{re}-initialization is performed before the training of subnetworks. At this time, both subnetworks achieves similar OOD performance to full network ERM. This ablation study confirms the importance of reusing initialization.

\textbf{Effects of bias relationship.} 
Our \textsc{FullColoredMNIST} keeps one fixed color-digit relationship for all biased data. Will rearranging this bias relationship defined in Section~\ref{sec:preliminaries} before the subnetwork is trained destroy the tickets? We then apply this to both winning tickets in Figure~\ref{fig:reseed_lth} and witness only a small accuracy drop of MRM after resetting the bias. This suggests our method indeed finds a subnetwork with a robust inductive bias for the invariant function, instead of only memorizing the bias relationship.

\begin{figure}[t]
    \hspace{0.1cm}
    \subfigure[\textsc{ColoredObject}]{
    \begin{minipage}{3.5cm}
    \includegraphics[width=3.5cm]{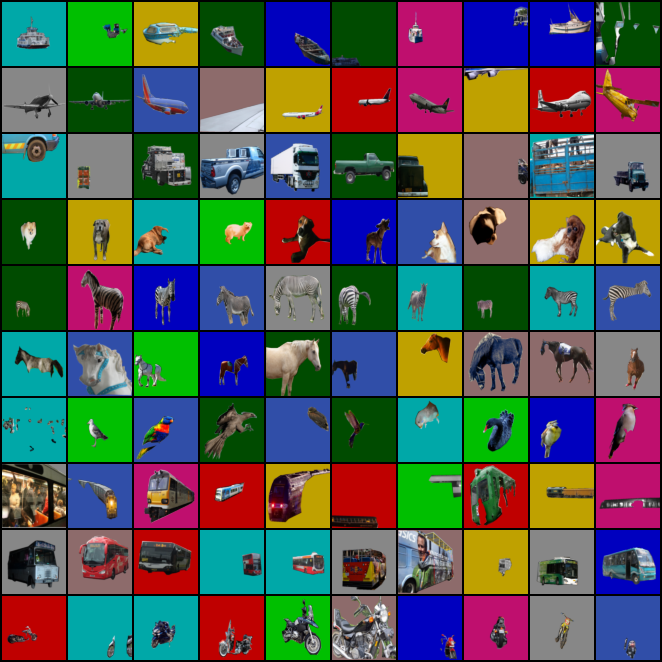}
    \end{minipage}
    \label{fig:coloredobject}
    }
    \hspace{0.1cm}
    \subfigure[\textsc{SceneObject}]{
    \begin{minipage}{3.5cm}
    \includegraphics[width=3.5cm]{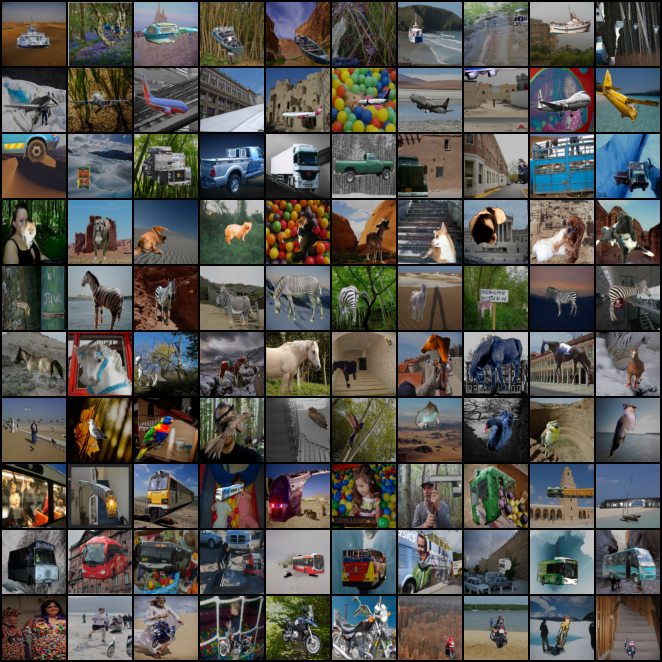}
    \end{minipage}
    \label{fig:sceneobject}
    }
    \caption{
    The visualization of \textsc{ColoredObject} (left) and \textsc{SceneObject} (right) datasets. We keep the same object (invariant feature) for each row and assign random backgrounds (spurious feature) to the images across different columns.
    }
    \label{fig:coco}
    \vspace{-0.45cm}
\end{figure}

\section{Experiments}
\label{sec: practical}

\begin{table}[t]
\caption{Generalization performance on \textsc{FullColoredMNIST}.}
\label{tab:fullcoloredmnist}
\vskip 0.15in
\begin{center}
\begin{small}
\begin{sc}
\begin{tabular}{c | cc}
\toprule
Methods & Train Accuracy & Test Accuracy  \\
\midrule \midrule
ERM & 98.10 \scriptsize{$\pm$ 0.09} & 57.75 \scriptsize{$\pm$ 1.84} \\
MRM & 98.90 \scriptsize{$\pm$ 0.05} & \textbf{72.98} \scriptsize{$\pm$ 0.58}  \\
\midrule
IRM & 98.18 \scriptsize{$\pm$ 0.09} & 59.30 \scriptsize{$\pm$ 1.88} \\
ModIRM & 98.77  \scriptsize{$\pm$ 0.12} & \textbf{70.86} \scriptsize{$\pm$ 2.12} \\
\midrule
REx & 98.86 \scriptsize{$\pm$ 0.10} & 75.61 \scriptsize{$\pm$ 1.26} \\
ModREx & 99.28  \scriptsize{$\pm$ 0.04} & \textbf{82.06} \scriptsize{$\pm$ 0.73} \\
\midrule
DRO   & 98.96 \scriptsize{$\pm$ 0.09} &  78.25 \scriptsize{$\pm$ 1.31} \\
ModDRO &99.39 \scriptsize{$\pm$ 0.04 } & \textbf{85.53} \scriptsize{$\pm$ 0.61}\\
\midrule
Unbias & 99.07 \scriptsize{$\pm$ 0.04 } & 99.03 \scriptsize{$\pm$ 0.08} \\
\bottomrule
\end{tabular}
\end{sc}
\end{small}
\end{center}
\end{table}

In this section, we demonstrate the effectiveness of our modular risk minimization algorithm on a variety of datasets. We compare our algorithm and its OOD variants with recent methods aiming at robust predictions across environments. For all methods we keep the same model architectures and training settings. We build three OOD datasets according to the bias protocol introduced in Section~\ref{sec:intro_OOD}: \textsc{FullColoredMNIST}, \textsc{ColoredObject} and \textsc{SceneObject}. For all datasets we design two training in-domains and one out-domain for evaluating OOD generalization capability. 
We defer all other experimental details to the supplementary materials.


\begin{table}[t]
\caption{Generalization performance on \textsc{ColoredObject}.}
\label{tab:coloredobject}
\vskip 0.15in
\begin{center}
\begin{small}
\begin{sc}
\begin{tabular}{c | cc}
\toprule
Methods & Train Accuracy & Test Accuracy  \\
\midrule \midrule
ERM & 87.56 \scriptsize{$\pm$ 2.52} & 43.74 \scriptsize{$\pm$ 2.11} \\
MRM  & 94.01 \scriptsize{$\pm$ 0.82} & \textbf{54.85} \scriptsize{$\pm$ 2.11}  \\
\midrule
IRM & 88.68 \scriptsize{$\pm$ 2.11} & 45.4  \scriptsize{$\pm$ 2.40} \\
ModIRM  & 93.01 \scriptsize{$\pm$ 0.36} & \textbf{52.35} \scriptsize{$\pm$ 1.28}  \\
\midrule
REx & 89.85 \scriptsize{$\pm$ 1.50} & 47.20 \scriptsize{$\pm$ 3.43} \\
ModREx & 93.55  \scriptsize{$\pm$ 1.45} & \textbf{55.51} \scriptsize{$\pm$ 2.76} \\
\midrule
DRO   & 91.73 \scriptsize{$\pm$ 0.40} &  51.95 \scriptsize{$\pm$ 1.62} \\
ModDRO & 92.67 \scriptsize{$\pm$ 0.92 } & \textbf{55.20} \scriptsize{$\pm$ 1.40}\\
\midrule
Unbias & 95.00 \scriptsize{$\pm$ 0.70 } & 72.37 \scriptsize{$\pm$ 2.53}\\
\bottomrule
\end{tabular}
\end{sc}
\end{small}
\end{center}

\vspace{-0.55cm}
\end{table}

\begin{table}[t]
\caption{Generalization performance on \textsc{SceneObject}.}
\label{tab:sceneobject}
\vskip 0.15in
\begin{center}
\begin{small}
\begin{sc}
\begin{tabular}{c | cc}
\toprule
Methods & Train Accuracy & Test Accuracy  \\
\midrule \midrule
ERM & 98.87 \scriptsize{$\pm$ 0.23} & 37.29 \scriptsize{$\pm$ 2.74} \\
MRM  & 99.61 \scriptsize{$\pm$ 0.04} & \textbf{39.44} \scriptsize{$\pm$ 0.77}  \\
\midrule
IRM & 98.68 \scriptsize{$\pm$ 0.27} & 37.19 \scriptsize{$\pm$ 2.58} \\
ModIRM  & 99.39 \scriptsize{$\pm$ 0.01} & \textbf{39.14} \scriptsize{$\pm$ 1.34}  \\
\midrule
REx & 92.91 \scriptsize{$\pm$ 1.11} & 38.84 \scriptsize{$\pm$ 1.39} \\
ModREx & 96.71  \scriptsize{$\pm$ 0.53} & \textbf{41.04} \scriptsize{$\pm$ 1.46} \\
\midrule
DRO   & 98.89 \scriptsize{$\pm$ 0.35} &  36.34 \scriptsize{$\pm$ 1.67} \\
ModDRO & 99.41 \scriptsize{$\pm$ 0.13 } & \textbf{39.14} \scriptsize{$\pm$ 1.60}\\
\midrule
Unbias & 95.25 \scriptsize{$\pm$ 2.21 } & 56.46 \scriptsize{$\pm$ 0.75} \\
\bottomrule
\end{tabular}
\end{sc}
\end{small}
\end{center}
\vspace{-0.5cm}
\end{table}

\textsc{FullColoredMNIST.} Details of the construction are in Section~\ref{sec:intro_OOD}. We summarize the results in Table~\ref{tab:fullcoloredmnist}.  We also use ERM trained with completely unbiased data whose bias coefficient is $(0.0, 0.0, 0.0)$ to serve as an upper bound (coined as ``Unbias" in the tables). 
Our method can consistently promote the OOD performance on this task, bringing around 10\% accuracy promotion.
The best behaved algorithm, ModDRO, reaches $85.53\%$ accuracy, contrary to the  78.25\% of top-grade baseline DRO  and  99.0\% achieved by unbiased solution.

\textsc{ColoredObject.}
We take inspiration from \citet{ahmed2021systematic} to build this biased dataset together with the following \textsc{SceneObject} one. Ten classes of objects extracted from MSCOCO dataset \cite{Lin2014Microsoft} are put onto ten kinds of color backgrounds. 
Figure~\ref{fig:coloredobject} displays 100 samples from this crafted biased dataset. Like \textsc{FullColoredMNIST}, we also set a one-to-one object-color relationship and set the bias coefficient differently as $(0.8, 0.6, 0.0)$. 
Results in Table~\ref{tab:coloredobject} demonstrate the advantages of our methods: all our four methods all achieve accuracy above 50\%, boosting their different baselines towards the optimal ``unbias"  solution.

\textsc{SceneObject.} 
Ten classes of objects extracted from MSCOCO dataset are put onto ten kinds of scenery backgrounds from Places dataset \cite{Zhou2018Places}. These scenery backgrounds make this task a more complex one than \textsc{ColoredObject}.  Figure~\ref{fig:sceneobject} displays 100 samples from this crafted dataset. Like \textsc{FullColoredMNIST}, we set a one-to-one object-scenery relationship and set the bias coefficient to be $(0.9, 0.7, 0.0)$, making it a even more biased and thus more difficult one than the previous task. This can also be shown with only 56.46\% accuracy of unbias solution. Corresponding results in Table~\ref{tab:sceneobject} shows that for this highly biased task, MRM and its variants can still accordingly improve out-distribution generalization performance in this highly bias setting, where previous OOD algorithms bring very limited benefit.

\vspace{-0.3cm}
\section{Related Work}
\vspace{-0.1cm}
\textbf{Out-of-distribution generalization.}
Machine learning beyond IID assumption is a very important problem and many research areas such as domain adaptation \cite{crammer2008learning, ben2010theory} and domain generalization \cite{muandet2013domain, motiian2017unified} have received much attention \cite{gulrajani2020search}. To get stable prediction for new unseen data distribution, it is desired to only rely on invariant features among the causal factorization of physical mechanisms of problem settings \cite{scholkopf2012causal}. \citet{peters2016causal} (ICP) claims that the residual of invariant method should remain IID and thus proposes to adopt statistical tests for mining invariant feature set. \citet{rojas2018invariant} generalizes this approach to nonlinear settings. 

Recently, since \citet{Arjovsky2019InvariantRM} brings invariant prediction into a more practical scenario, a large amount of works has made solid progress for alleviating spurious correlation and shortcut exploitation \cite{geirhos2020shortcut, koh2020wilds}: \citet{sagawa2019distributionally} proposes to use group DRO when attribution information is provided; \citet{chang2020invariant} incorporates this invariant inference idea into selective rationalization area; \citet{ahuja2020invariant} studies the IRM formulation from a game theory and bilevel optimization formulation; \citet{Krueger2020OutofDistributionGV} propose REx to enforce the variance of losses across distribution, which is further analyzed by \citet{xie2020risk}; \citet{koyama2020out} (IGA) also has a similar contribution with different theoretical analysis;  \citet{jin2020domain} (RGM) proposes another training objective from regret minimization viewpoint;
\citet{pezeshki2020gradient} studies the gradient starvation phenomenon which is connected with spurious correlation and proposes an insightful solution;
\citet{creager2020exchanging} (EIIL) points out that invariant prediction shares the same spirit with fair representation learning; \citet{parascandolo2020learning} (ILC) proposes to focus second order landscape information;
\citet{ahmed2021systematic} adopts a divergence term to match the output distribution spaces of different domains;
\citet{muller2020learning} achieves invariance from an information theory start point and enforces conditional invariance with HSIC terms.
Some other works also point out the pitfalls of current approaches, showing only in very limited situations can \citet{Arjovsky2019InvariantRM} (\textit{e.g.}, low dimension settings) really capture invariance: \citet{rosenfeld2020risks} proves the validity of IRM for linear cases but gives a negative example for nonlinear cases; \citet{nagarajan2020understanding} analyzes different failure modes of OOD generalization; \citet{ahuja2020empirical} analyze the sample efficiency properties of IRM; \citet{kamath2021does} investigates the success and failure cases of IRM and IRMv1 on simple but insightful settings, and claims the community might need a better invariance notion.

Another line of works study a related but different topic named \textit{debiasing}, where there is no explicit multiple environments setting provided. Bias in realistic datasets are usually exploited in a spurious way, such as the texture-bias of Imagenet-trained models \cite{geirhos2018imagenet}. Subsequent works \cite{wang2019learning, bahng2020learning, shi2020informative, nam2020learning, li2021shapetexture, sauer2021counterfactual} focus on addressing the bias problem with explicit debiasing procedure.

\textbf{Modularity.}
Modularity \cite{ballard1987modular, fodor1988connectionism, newman2006modularity} has been considered as a crucial part of intelligent systems. 
Lots of works focus on imposing explicit module level modularity \cite{clune2013evolutionary, andreas2016neural, chang2018automatically, goyal2021recurrent}, while others also explore weight level modularity in a more fine-grained way \cite{mallya2018packnet, watanabe2019understanding, filan2020neural, csordas2020neural}. Our work also belongs to the latter category.

\textbf{Pruning.} 
We mainly focus on unstructured pruning literature. 
This line of model compression literature dates back to \citet{1989Skeletonization, lecun1989optimal, hassibi1993second} with more recent pruning methods \cite{han2015deep, molchanov2016pruning, dong2017learning}. Recently, the lottery ticket hypothesis \cite{frankle2018lottery} sheds more light into this field, showing the importance of initialization. \cite{liu2018rethinking} also propose another viewpoint that for practical settings the inherited weights are not important.

\vspace{-0.3cm}
\section{Discussion}

\textbf{Data settings.} 
The seminal work \cite{Arjovsky2019InvariantRM} proposes to use color in digit identification as a spurious correlation.
In order to exposit the effectiveness of IRM, the authors enforce a 25\% label noise in the binary classification data and assign color a \textit{larger} correlation than the true digit shape. In this way, ERM exploits color feature to predict. 
While there is controversy surrounding whether one should still treat digit as desired learning target under this situation, we choose to impose no label noise in \textsc{FullColoredMNIST} as is the case in \citet{nam2020learning, ahmed2021systematic}. This choice enables the structure learning procedure could mine the true invariant feature.
On the other hand, our work is limited as we haven't considered the data settings such as group attribution available ones \cite{sagawa2019distributionally, xie2020n, khani2021removing} and we shall fill this gap in future work. More about datasets can be found in Section~\ref{sec:data_detail}.

\textbf{Success of MRM.} 
There are several reasons for why MRM can improve OOD performance without invariance constrain. 
The first reason is related to our label noise free setting discussed above.
This makes the invariant feature itself perfectly predictive of the label, thus containing all information about the desired target function. Then the problem would be how to exploit this information effectively. 
MRM becomes competent for OOD tasks by providing a novel and helpful parameterization method for the original optimization problem with extra parameters.
One notable thing is that more structure parameters actually don't increase the expressive power of the neural network, since every weight can take zero value in nature.
Another reason is we adopt an explicit approach to zero out the ``spurious part" of the model weights, hence achieving a not-so-biased solution. 
Notice this cannot be achieved with random sparse model,  revealing the structure is a key element for OOD generalization.
Therefore, a positive answer is given to the title of this work.
We further refer to Section~\ref{sec:mask_ablation} for empirical results of the importance of a proper sparsity level in structure learning.





\vspace{-0.3cm}
\section*{Acknowledgement}
\vspace{-0.2cm}
Kartik Ahuja acknowledges the support provided by IVADO postdoctoral fellowship funding program. Yilun Xu is supported by the MIT HDTV Grand Alliance Fellowship. Yisen Wang is partially supported by the National Natural Science Foundation of China under Grant 62006153, and CCF-Baidu Open Fund (OF2020002). Aaron Courville acknowledges the funding from CIFAR Canadian AI Chair and Hitachi. The authors would also like to thank Róbert Csordás, David Krueger, Faruk Ahmed, Mohammad Pezeshki,  Baifeng Shi, Sara Hooker and anonymous reviewers for insightful discussion and feedbacks. 

\newpage
\bibliography{bib}
\bibliographystyle{icml2021}


\onecolumn
\appendix

\section{Omitted Proof}
\subsection{Proof for Proposition~\ref{prop:linear}}

\begin{proof}
We first analyze the performance of $f_{\mathsf{reg}}$.  The prediction from this classifier is $\hat{Y}^e = \mathsf{sgn}(\boldsymbol{w}_{\mathsf{spu}}^{\mathsf{T}}Z_{\mathsf{sp}}^e)$, thus

\begin{equation}
    \begin{split}
         \boldsymbol{w}_{\mathsf{spu}}^{\mathsf{T}}Z_{\mathsf{sp}}^e &= \frac{1}{\sqrt{D}}\sum_{i=1}^{D}Z_{\mathsf{sp},i}^e = \sqrt{D}\Big[ \frac{1}{D}\sum_{i=1}^{D}Z_{\mathsf{sp},i}^e\Big], \\
         \hat{Y}^e &= \mathsf{sgn} (\boldsymbol{w}_{\mathsf{spu}}^{\mathsf{T}}Z_{\mathsf{sp}}^e) = \mathsf{sgn}\Big[ \frac{1}{D}\sum_{i=1}^{D}Z_{\mathsf{sp},i}^e\Big]. 
    \end{split}
    \label{proof:eqn1}
\end{equation}

We then analyze the error for any environment:
\begin{equation}
\begin{split}
     \mathsf{Err}^{e} &= \frac{1}{2} \Big[ 1 - \mathbb{E}^{e}\big[\hat{Y}^eY^e\big]\Big], \\ 
     \mathbb{E}^{e}\big[\hat{Y}^eY^e\big] &=  \mathbb{E}^{e}\Big[\mathsf{sgn}\Big[ \frac{1}{D}\sum_{i=1}^{D}Z_{\mathsf{sp},i}^e\Big]  Y^e\Big] 
     =\sum_{y\in \{-1,1\}} \mathbb{P}[Y^{e}=y]\mathbb{E}^{e}\Big[\mathsf{sgn}\Big( \frac{1}{D}\sum_{i=1}^{D}Z_{\mathsf{sp},i}^e\Big)|Y^e=y\Big]y ,
\end{split} 
    \label{proof:eqn2}
\end{equation}

\begin{equation}
    \begin{split}
         \mathbb{E}^{e}\Big[\mathsf{sgn}\Big( \frac{1}{D}\sum_{i=1}^{D}Z_{\mathsf{sp},i}^e\Big)|Y^e=1\Big] &=  
         \mathbb{P}\Big[\frac{1}{D}\sum_{i=1}^{D}Z_{\mathsf{sp},i}^e>0|Y^e=1\Big] - \mathbb{P}\Big[\frac{1}{D}\sum_{i=1}^{D}Z_{\mathsf{sp},i}^e\leq 0|Y^e=1\Big] \\ 
        &= 2\mathbb{P}\Big[\frac{1}{D}\sum_{i=1}^{D}Z_{\mathsf{sp},i}^e>0|Y^e=1\Big]-1.
    \end{split}
    \label{proof:eqn3}
\end{equation}

Observe that $\mathbb{P}\Big[\frac{1}{D}\sum_{i=1}^{D}Z_{\mathsf{sp},i}^e<0|Y^e=1\Big] = \mathbb{P}\Big[\frac{1}{D}\sum_{i=1}^{D}Z_{\mathsf{sp},i}^e>0|Y^e=-1\Big]$. Using this observation and plugging  \eqref{proof:eqn3} into \eqref{proof:eqn2} we get 
\begin{equation}
    \begin{split}
        \mathsf{Err}^{e} = \mathbb{P}\Big[\frac{1}{D}\sum_{i=1}^{D}Z_{\mathsf{sp},i}^e\leq 0 |Y^e=1\Big].
    \end{split}
\end{equation}

Let us now bound $\mathsf{Err}^{e}$.
Define $\bar{Z}_{\mathsf{sp}}^e = \frac{1}{D}\sum_{i=1}^{D}Z_{\mathsf{sp},i}^e$. 
Since 
\begin{equation}
    \mathbb{E}^e\Big[\frac{1}{D}\sum_{i=1}^{D}Z_{\mathsf{sp},i}^e|Y^e=1\Big] = 2p^e-1,
\end{equation}
we have
\begin{equation}
    \begin{split}
        \mathbb{P}\Big[\frac{1}{D}\sum_{i=1}^{D}Z_{\mathsf{sp},i}^e\leq 0|Y^e=1\Big] &= \mathbb{P}\Big[\bar{Z}_{\mathsf{sp}}\leq 0|Y^e=1\Big] \\ 
        & =\mathbb{P}\Big[\bar{Z}_{\mathsf{sp}}^e-\mathbb{E}[\bar{Z}_{\mathsf{sp}}^e] \leq -\mathbb{E}[\bar{Z}_{\mathsf{sp}}^e]|Y^e=1\Big] 
         \leq \mathbb{P}\Big[|\bar{Z}_{\mathsf{sp}}^e-\mathbb{E}[\bar{Z}_{\mathsf{sp}}^e]|\geq \mathbb{E}[\bar{Z}_{\mathsf{sp}}^e]|Y^e=1\Big] \\
        & \leq 2e^{-2(2p^e-1)^2D} \leq 2e^{-2c^2D}.
    \end{split}
\end{equation}

In the test environment, since $Z^e$ and $Y^e$ are independent and $p^e=0.5$. As a result, the error in test environment for the regular classifier is          $\mathsf{Err}^{e} = \mathbb{P}\Big[\frac{1}{D}\sum_{i=1}^{D}Z_{\mathsf{sp},i}^e\leq 0 |Y^e=1\Big] = \mathbb{P}\Big[\frac{1}{D}\sum_{i=1}^{D}Z_{\mathsf{sp},i}^e\leq 0 \Big] =  0.5$. 

Now let us consider the optimal sparse max-margin classifier. For $d=2$, the max-margin classifier for the above data distribution is simply $w_{\mathsf{inv}}=1$ and $\boldsymbol{w}_{\mathsf{sp}}=\boldsymbol{0}$. Since $Z_{\mathsf{inv}}^e=Y^e$, in both train and test, the sparse classifier has a perfect accuracy in both train and test environments. 

Next, we compare the margins. The margin for $f_{\mathsf{reg}}$ is $Y^e \boldsymbol{w}_{\mathsf{sp}}^{\mathsf{T}}Z_{\mathsf{sp}} \geq \frac{c\sqrt{D}}{2}$ with a probability at least $1-\delta$ (the proof follows directly from Hoeffding's inequality and we refer to the  Appendix A of \citet{nagarajan2020understanding}). In comparison, $f_{\mathsf{sparse}}^{d}$ that assigns weights to invariant and spurious parts as follows $w_{\mathsf{inv}}=1$ and $\boldsymbol{w}_{\mathsf{sp}}=\boldsymbol{0}$ achieves a margin of $Y^e w_{\mathsf{inv}}Z_{\mathsf{inv}} =  1$.

\end{proof}

\section{More Discussion}

One possible algorithm we do not explore in this work is the IRM games \cite{ahuja2020invariant}.
This approach sees the IRM formulation from a game theory perspective across different environments and design a corresponding algorithm.
The algorithm has one network per environment thus the number of parameters scale in number of environments. 
In order to make comparisons apple to apple, we keep all the methods to have the same parameter complexity. IRM games will have more parameters so we do not compare with it. Other methods including extra parameters (\textit{e.g.}, auxiliary neural networks) are also not considered due to analogous reason and we shall explore them in future work.

We notice there is a deep connection between \citet{parascandolo2020learning} and our work. Our MRM explicitly learn a subnetwork architecture and only update the corresponding subset of weights, while \citet{parascandolo2020learning} also restricts its optimization within a subset of weights by ``and mask" algorithm, which only updates a parameter when gradients across domains are consensus to each other. However, this method may need a appropriate large number of domains to work (see details in their paper) and hence is not among the studied algorithms in this paper. 
In the future work, we intend to explore the relationship between the structure of our digit module and the subset found by their ``and mask" under more domains.

Similar to our method, Dropout \cite{srivastava2014dropout} also only updates part of the network during training. We list some of the main difference here: Dropout aims to prevent overfitting by simply not updating whole model parameters towards one single function, while we pursue to identify one particular functional module architecture within full network; Dropout only randomly zero out the updating gradients, while we intentionally pick particular functional part of the model in an end-to-end way; what's more, Dropout is not activated during testing inference time, while our subnetwork is kept for all following stages.

In \citet{hooker2019compressed, hooker2020characterising}, the authors propose that model compression will hurt the accuracy on underrepresented groups with negligible impacts on overall accuracy, which seems contradictory to our results. Here we state about the difference in the settings and claim that their works are actually consistent in spirit to ours. First of all, their work do not target OOD problems or invariant prediction, but focus on long tail underrepresented subgroups in IID situations.
In our context ``bias" means a spurious / shortcut way of inference and we aim to zero out the spurious part in the model parameters, while their settings don't contain a spurious feature that will bias the model prediction, hence most of the parameters are rightful and shouldn't be got rid of. As a result, it's natural for model compression to hurt in their cases.  What's more, our method utilize a much more careful subnetwork selection method where we aim to maximize the in-domain performance when searching structures, ensuring that we do not wipe off the useful and rightful part of parameters. 

 In \citet{bengio2019meta} and its following analysis \cite{priol2020analysis}, the authors claim that a better knowledge of true causal relationship will bring a faster speed of transferring learning. While on the other hand, from Figure~\ref{fig:erm_lth} we can see a good subnetwork leads to faster convergence and better performance. Although we do not target a \textit{causally} promising algorithm in this work (our main goal is to get rid of spurious correlation for OOD problems), we claim our approach is connected with causal discovery \cite{bengio2019meta} stated above, in the sense of capturing causal relationship hidden in data better (please note that invariant prediction can be seen as a higher level of transfer learning in an OOD sense).

\section{More on Experiments}

\subsection{Dataset details}
\label{sec:data_detail}
\textsc{FullColoredMNIST.} We use ten colors taken from \citet{ahmed2021systematic} for all the data. Their RGB values are: [0, 100, 0], [188, 143, 143], [255, 0, 0], [255, 215, 0], [0, 255, 0], [65, 105, 225], [0, 225, 225], [0, 0, 255], [255, 20, 147], [160, 160, 160]. Each image is of size $3\times 32\times 32$ and the whole dataset contains $60000$ images.
The difference between \citet{ahmed2021systematic} and ours are significant: their whole data is deemed to come from ``majority group" and ``minority group", where the images from ``majority group" are colored with previous mentioned ten colors and those from ``minority group" are colored with \textit{other fifty different colors}. In contrast, all of our images are colored with these ten colors. We set a bias relationship to connect each digit and each color one by one and define  bias coefficient to be the ratio of bias data that follows the relationship, as described in Section~\ref{sec:preliminaries}.
Besides, their data are pooled and integrated as one domain before presented to algorithms. Our dataset is also different with \citet{Arjovsky2019InvariantRM}, where they use label noise intentionally to make the correlation between label and color to be higher than the one between label and digit. This makes ERM severely biased towards color and thus fails in test domain. On the contrary, we do not impose any sort of label noise and leave label as the ground truth digit information. 
What's more, \citet{Arjovsky2019InvariantRM} use binary classification and two colors, which is a much simpler setting. Our setting is also different from \citet{nam2020learning}, which is not  multi environment and only has one domain.

\textsc{ColoredObject.} We imitate \citet{ahmed2021systematic} to take ten objects as invariant features and put them on ten different color backgrounds. We use the ten colors mentioned above, and take the following ten objects: boat, airplane, truck, dog, zebra, horse, bird, train, bus, motorcycle. Each image is of size $3\times 64\times 64$ and the whole dataset contains $10000$ images. The difference between ours and \citet{ahmed2021systematic} is similar to \textsc{FullColoredMNIST}: their settings contain a minority group which has many other backgrounds than the mentioned ten color backgrounds.

\textsc{SceneObject.} We imitate \citet{ahmed2021systematic} to take ten objects as invariant features and put them on ten different scenery backgrounds (beach, canyon, building facade, staircase, desert sand, crevasse, bamboo forest, broadleaf, ball pit and kasbah). We use the same ten object classes mentioned above.
Each image is of size $3\times 64\times 64$ and the whole dataset contains $10000$ images. The difference between ours and \citet{ahmed2021systematic} is similar to \textsc{FullColoredMNIST}: their settings contain a minority group which has many other backgrounds than the mentioned ten scenery backgrounds.

\begin{table}[t]
\caption{Generalization performance on \textsc{FullColoredMNIST} with oracle validation.}
\vskip 0.15in
\begin{center}
\begin{small}
\begin{sc}
\label{tab:oracle_coloremnist}
\begin{tabular}{c | cc}
\toprule
Methods & Train Accuracy & Test Accuracy  \\
\midrule \midrule
ERM & 98.10 \scriptsize{$\pm$ 0.10} & 58.04 \scriptsize{$\pm$ 1.95} \\
MRM & 98.90 \scriptsize{$\pm$ 0.05} & \textbf{73.21} \scriptsize{$\pm$ 0.58}  \\
\midrule
IRM & 98.17 \scriptsize{$\pm$ 0.12} & 59.55 \scriptsize{$\pm$ 1.90} \\
ModIRM & 98.67  \scriptsize{$\pm$ 0.20} & \textbf{70.35} \scriptsize{$\pm$ 3.22} \\
\midrule
REx & 98.83 \scriptsize{$\pm$ 0.09} & 76.17 \scriptsize{$\pm$ 1.53} \\
ModREx & 99.28  \scriptsize{$\pm$ 0.05} & \textbf{82.13} \scriptsize{$\pm$ 0.82} \\
\midrule
DRO   & 98.94 \scriptsize{$\pm$ 0.11} &  78.56 \scriptsize{$\pm$ 1.42} \\
ModDRO &99.38 \scriptsize{$\pm$ 0.06 } & \textbf{85.67} \scriptsize{$\pm$ 0.51}\\
\midrule
Unbias & 99.05 \scriptsize{$\pm$ 0.04 } & 97.86 \scriptsize{$\pm$ 0.20  } \\
\bottomrule
\end{tabular}
\end{sc}
\end{small}
\end{center}
\end{table}

\begin{table}[t]
\caption{Generalization performance on \textsc{ColoredObject} with oracle validation.}
\label{tab:oracle_coloredobject}
\vskip 0.15in
\begin{center}
\begin{small}
\begin{sc}
\begin{tabular}{c | cc}
\toprule
Methods & Train Accuracy & Test Accuracy  \\
\midrule \midrule
ERM & 87.58 \scriptsize{$\pm$ 2.42} & 44.39 \scriptsize{$\pm$ 2.44} \\
MRM  & 94.00 \scriptsize{$\pm$ 0.56} & \textbf{55.03} \scriptsize{$\pm$ 1.76}  \\
\midrule
IRM & 87.63 \scriptsize{$\pm$ 2.32} & 44.49 \scriptsize{$\pm$ 2.15} \\
ModIRM  & 92.95 \scriptsize{$\pm$ 0.41} & \textbf{52.55} \scriptsize{$\pm$ 0.65}  \\
\midrule
REx & 89.28 \scriptsize{$\pm$ 1.35} & 46.07 \scriptsize{$\pm$ 2.59} \\
ModREx & 82.65  \scriptsize{$\pm$ 1.85} & \textbf{55.56} \scriptsize{$\pm$ 3.16} \\
\midrule
DRO   & 91.84 \scriptsize{$\pm$ 2.17} &  53.20 \scriptsize{$\pm$ 1.15} \\
ModDRO & 92.32 \scriptsize{$\pm$ 1.37 } & \textbf{55.46} \scriptsize{$\pm$ 1.43}\\
\midrule
Unbias & 92.32 \scriptsize{$\pm$ 1.80 } & 72.77 \scriptsize{$\pm$ 3.48}\\
\bottomrule
\end{tabular}
\end{sc}
\end{small}
\end{center}
\end{table}

\begin{table}[th]
\caption{Generalization performance on \textsc{SceneObject} with oracle validation.}
\label{tab:oracle_sceneobject}
\vskip 0.15in
\begin{center}
\begin{small}
\begin{sc}
\begin{tabular}{c | cc}
\toprule
Methods & Train Accuracy & Test Accuracy  \\
\midrule \midrule
ERM & 95.04 \scriptsize{$\pm$ 2.13} & 37.44 \scriptsize{$\pm$ 1.15} \\
MRM  & 98.83 \scriptsize{$\pm$ 0.26} & \textbf{39.49} \scriptsize{$\pm$ 0.15}  \\
\midrule
IRM & 92.66 \scriptsize{$\pm$ 0.24} & 37.54 \scriptsize{$\pm$ 0.94} \\
ModIRM  & 94.63 \scriptsize{$\pm$ 0.54} & \textbf{40.04} \scriptsize{$\pm$ 1.90}  \\
\midrule
REx & 92.73 \scriptsize{$\pm$ 1.61} & 39.39 \scriptsize{$\pm$ 0.96} \\
ModREx & 96.72  \scriptsize{$\pm$ 0.53} & \textbf{40.79} \scriptsize{$\pm$ 1.62} \\
\midrule
DRO   & 94.53 \scriptsize{$\pm$ 2.17} &  36.64 \scriptsize{$\pm$ 1.35} \\
ModDRO & 93.52 \scriptsize{$\pm$ 5.90 } & \textbf{40.99} \scriptsize{$\pm$ 1.70}\\
\midrule
Unbias & 85.47 \scriptsize{$\pm$ 2.37 } & 56.91 \scriptsize{$\pm$ 1.31} \\
\bottomrule
\end{tabular}
\end{sc}
\end{small}
\end{center}
\end{table}

\subsection{Experimental details}
For the results in the main text, we take a commonly used policy to report the last step accuracy. We also apply the another evaluation method (the ``oracle validation" in \citet{gulrajani2020search}) to report accuracy, and provide corresponding results in Table~\ref{tab:oracle_coloremnist}, \ref{tab:oracle_coloredobject} and \ref{tab:oracle_sceneobject}. These results are consistent to those in main text, showing the validity of our conclusion.
For both method, we take a similar approach to \citet{gulrajani2020search}, and the difference is that we take a finite search set for hyperparameters instead of sampling of a human defined distribution.
For all datasets, we search the regularization coefficient of IRM and REx in $\{1e-1, 5e-1, 1, 1e1, 1e2, 1e3, 1e4\}$, the step when the regularization is added into training in $\{0, 1000, 2000\}$. Furthermore, we also search a binary option about whether to scale down the whole loss term by the regularization coefficient as in \citet{Arjovsky2019InvariantRM}. For DRO we search the group proportion step size $\eta_q$ (notation taken from \citet{sagawa2019distributionally}) in $\{1e-4, 1e-3, 1e-2, 1e-1, 1\}$. For the subnetwork structure learning, we follow \citet{csordas2020neural} to use Adam optimizer and search the logit learning rate among $\{1e-2, 1e-1, 1\}$ and the sparsity coefficient among $\{1e-8, 1e-7, 1e-6, 1e-5, 1e-4, 1e-3\}$.
For Figure~\ref{fig:lth_sparsity}, the x-axis is in log scale and ranges from $1e-5$ to $1e-3$ for sparse cases.

For \textsc{FullColoredMNIST}, 
all experiments are measured by computing mean and standard deviation across five trials with random seeds. For optimization we use SGD + momentum (0.9) with $1e-4$ weight decay, where the initial learning rate is $1e-1$ and is decayed every $600$ steps for all algorithms. We take the batch size to be $128$. The training process longs for $2000$ steps (and thus $N_1$ is also $2000$). $N_2$ is set to $2000$, and it can be seen from Table~\ref{tab:n2} that the importance of this hyperparameter is very limited. 
We use a simple ConvNet with three convolutional layers with feature map dimensions of 64, 128 and 256, each followed by a ReLU nonlinear and a batch normalization layer. The fourth layer is a fully connected layer.
For \textsc{ColoredObject}, all experiments are measured by computing mean and standard deviation a cross three trials with random seeds. The learning rate is $1e-1$ and decays every $1200$ steps. The training process longs for $2000$ steps. Others are kept the same with previous dataset.  
We use Wide ResNet 28-2 architecture \cite{zagoruyko2016wide}. To adapt to $64\times 64$ image size, we replace the average pooling layer with window size $8$ to one with size $16$.
For \textsc{SceneObject}, we use a learning rate of $5e-2$ which decays every $1200$ steps. Other unmentioned settings are kept consistent as above.
We use Tesla V100 GPU to perform the experiments.

\subsection{More empirical results}

\begin{figure}[th]
    \subfigure[Convergence speed for ERM.]{
    \begin{minipage}{4cm}
    \includegraphics[width=4cm]{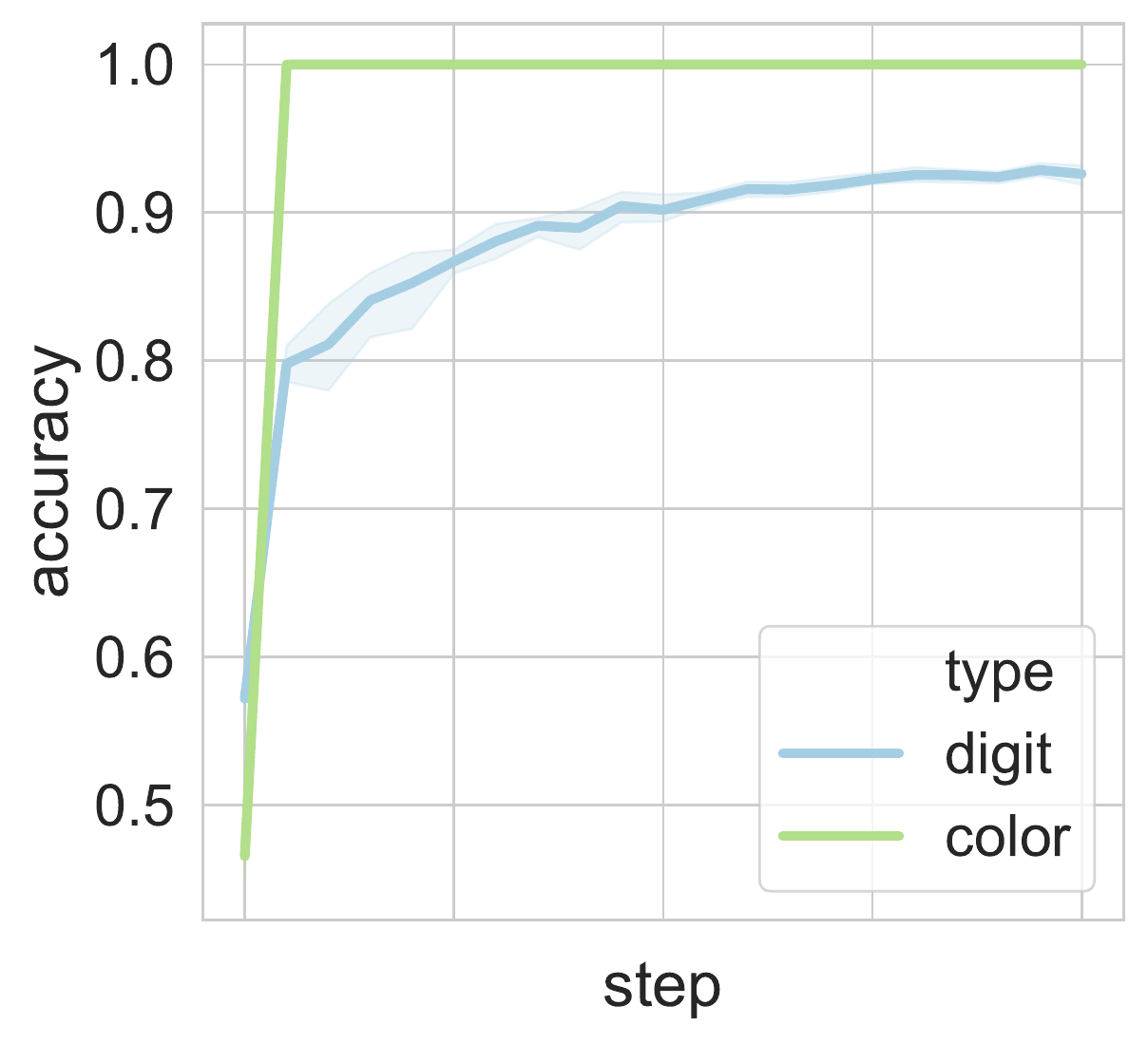}
    \end{minipage}
    \label{fig:module_convergence}
    }
    \subfigure[Accuracy of module for IRM.]{
    \begin{minipage}{4cm}
    \includegraphics[width=4cm]{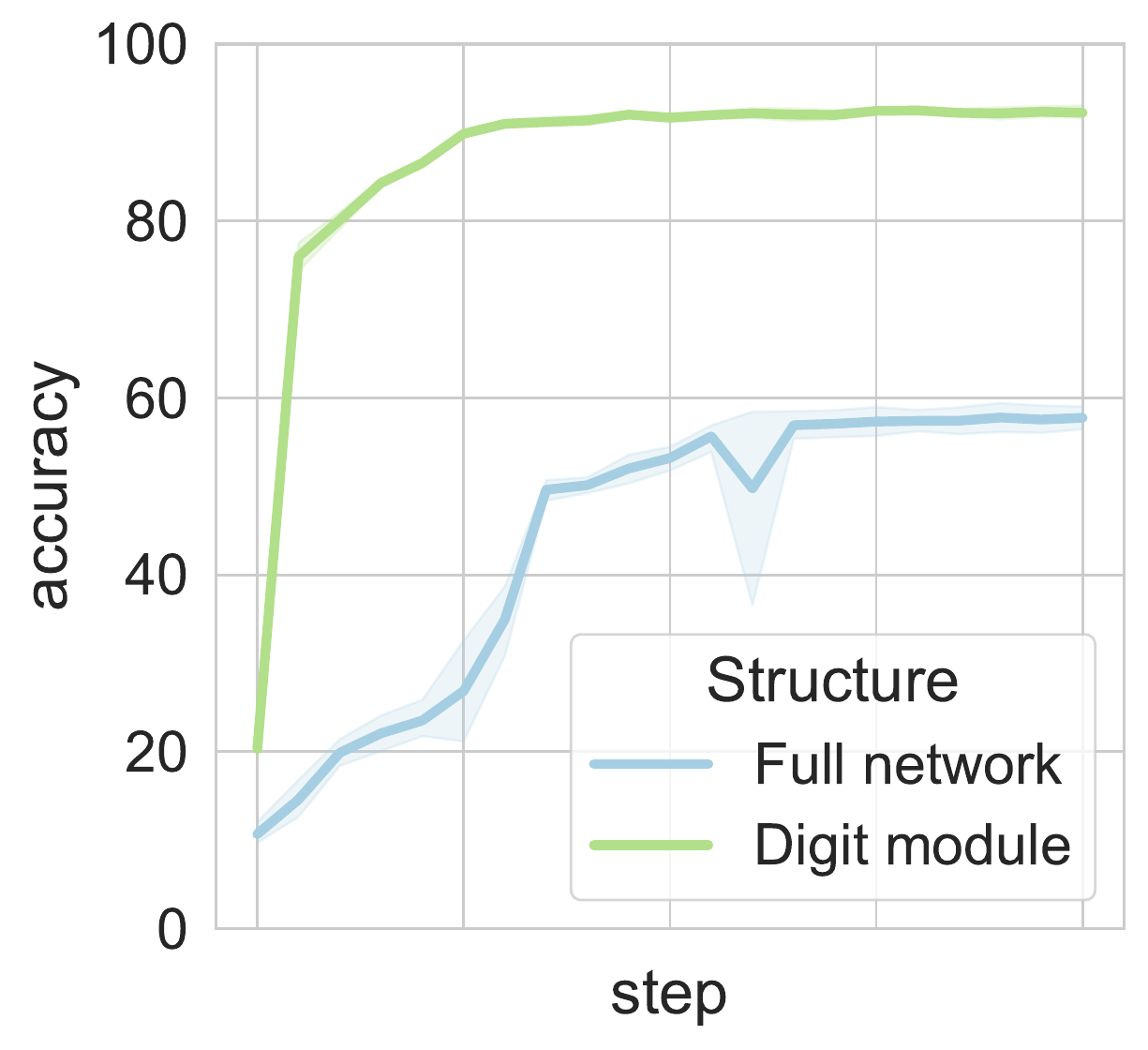}
    \end{minipage}
    \label{fig:irm_digit_acc}
    }
    \subfigure[Accuracy of module for REx.]{
    \begin{minipage}{4cm}
    \includegraphics[width=4cm]{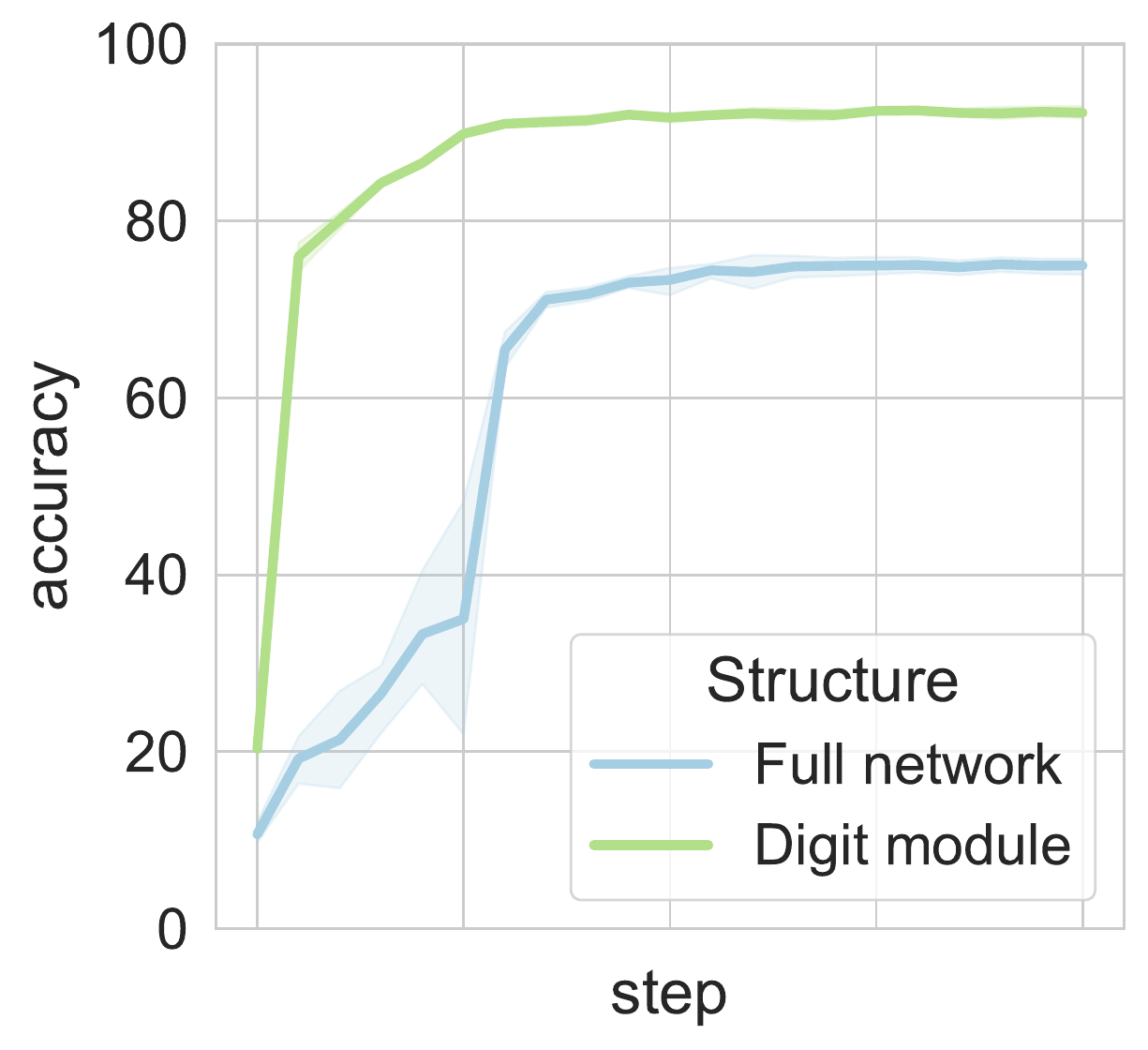}
    \end{minipage}
    \label{fig:rex_digit_acc}
    }
    \subfigure[Accuracy of module for DRO.]{
    \begin{minipage}{4cm}
    \includegraphics[width=4cm]{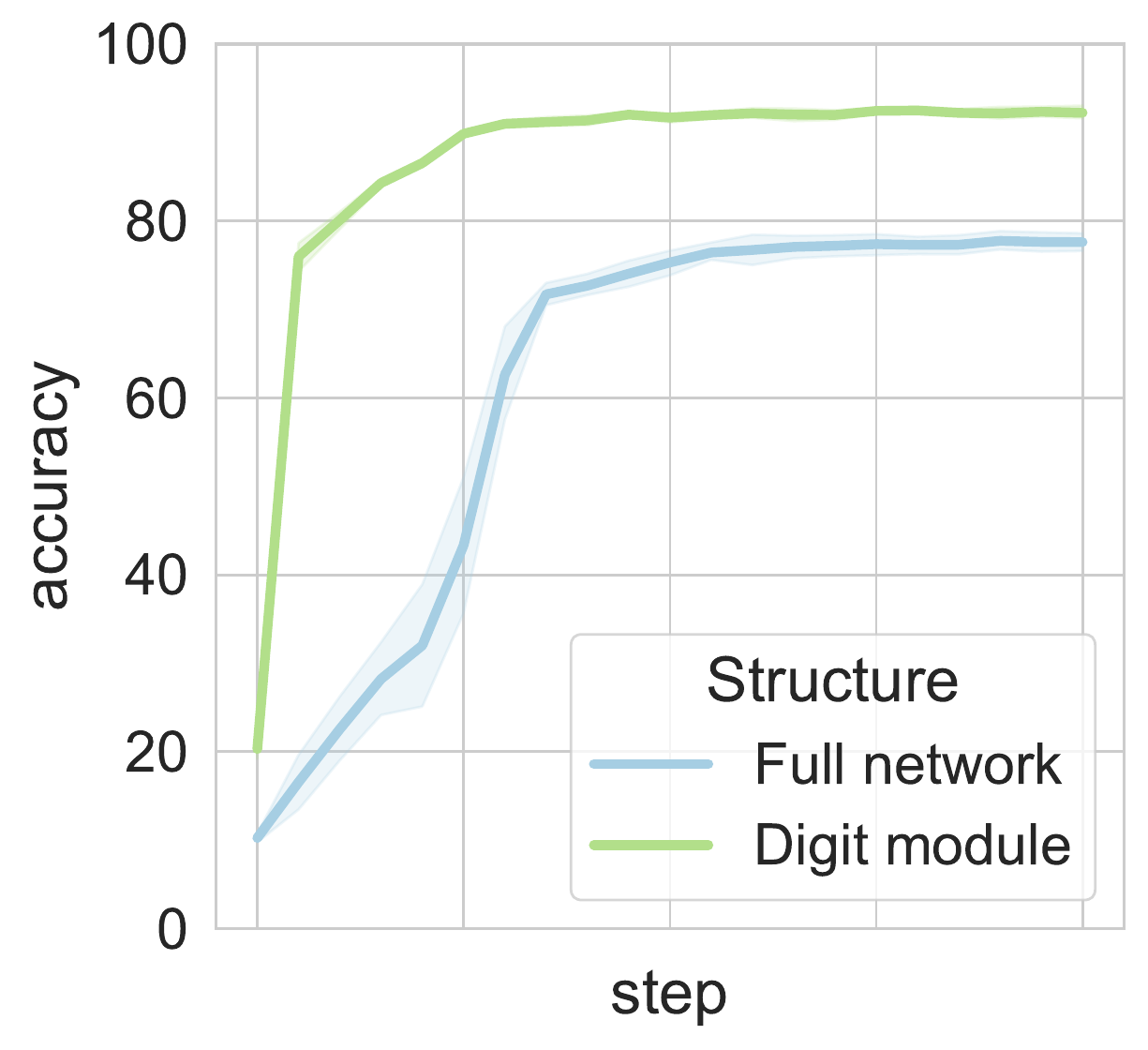}
    \end{minipage}
    \label{fig:dro_digit_acc}
    }
    \caption{(a) Convergence speed comparison for modules of ERM. (b)(c)(d) Analogy of Figure~\ref{fig:digit_acc} for other algorithms.}
\end{figure}


We affirm that color is more fitted to neural network prior by plotting the digit accuracy of digit module and color accuracy of color module w.r.t. the training process in Figure~\ref{fig:module_convergence}. The two modules are both learned given the same ERM trained model at every step. The result shows that the training of color module converges much faster than the digit one, and thus showing that neural networks have a natural favor for color (texture) information than digit (shape) information as shown in \citet{zeiler2014visualizing, ritter2017cognitive, brendel2019approximating, shi2020informative}.
We also plot the behaviors of digit module learning of three OOD algorithms in Figure~\ref{fig:irm_digit_acc} \ref{fig:rex_digit_acc} and \ref{fig:dro_digit_acc}. These three plots show a very similar results to Figure~\ref{fig:digit_acc}.
We omit the behavior of color module of these algorithms since it's very similar to that of ERM shown in Figure~\ref{fig:module_convergence}.

\begin{figure}[t]
    \subfigure[Intersection, digit accuracy.]{
    \begin{minipage}{4cm}
    \includegraphics[width=4cm]{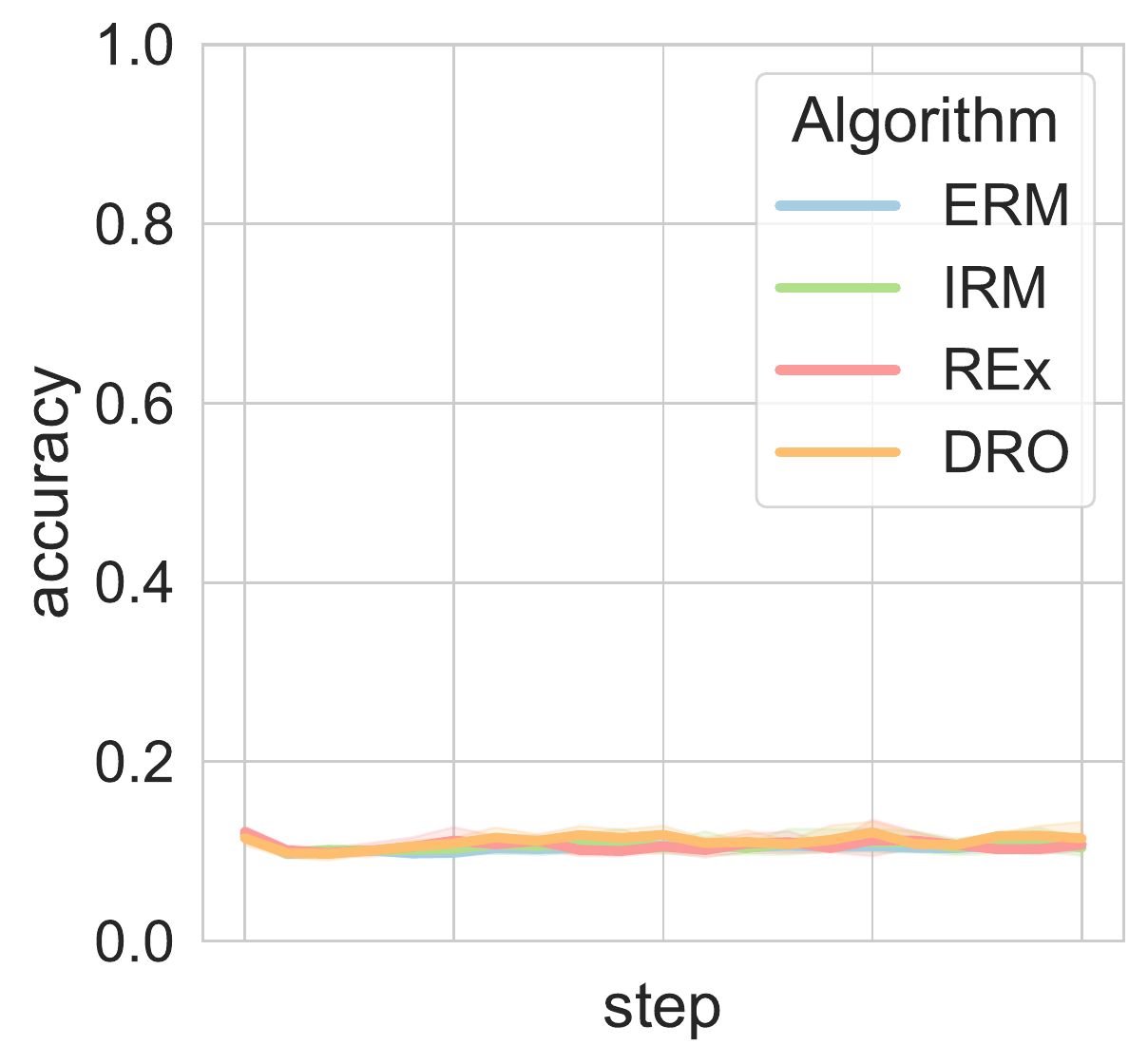}
    \end{minipage}
    \label{fig:and_mask_acc}
    }
    \subfigure[Intersection, color accuracy.]{
    \begin{minipage}{4cm}
    \includegraphics[width=4cm]{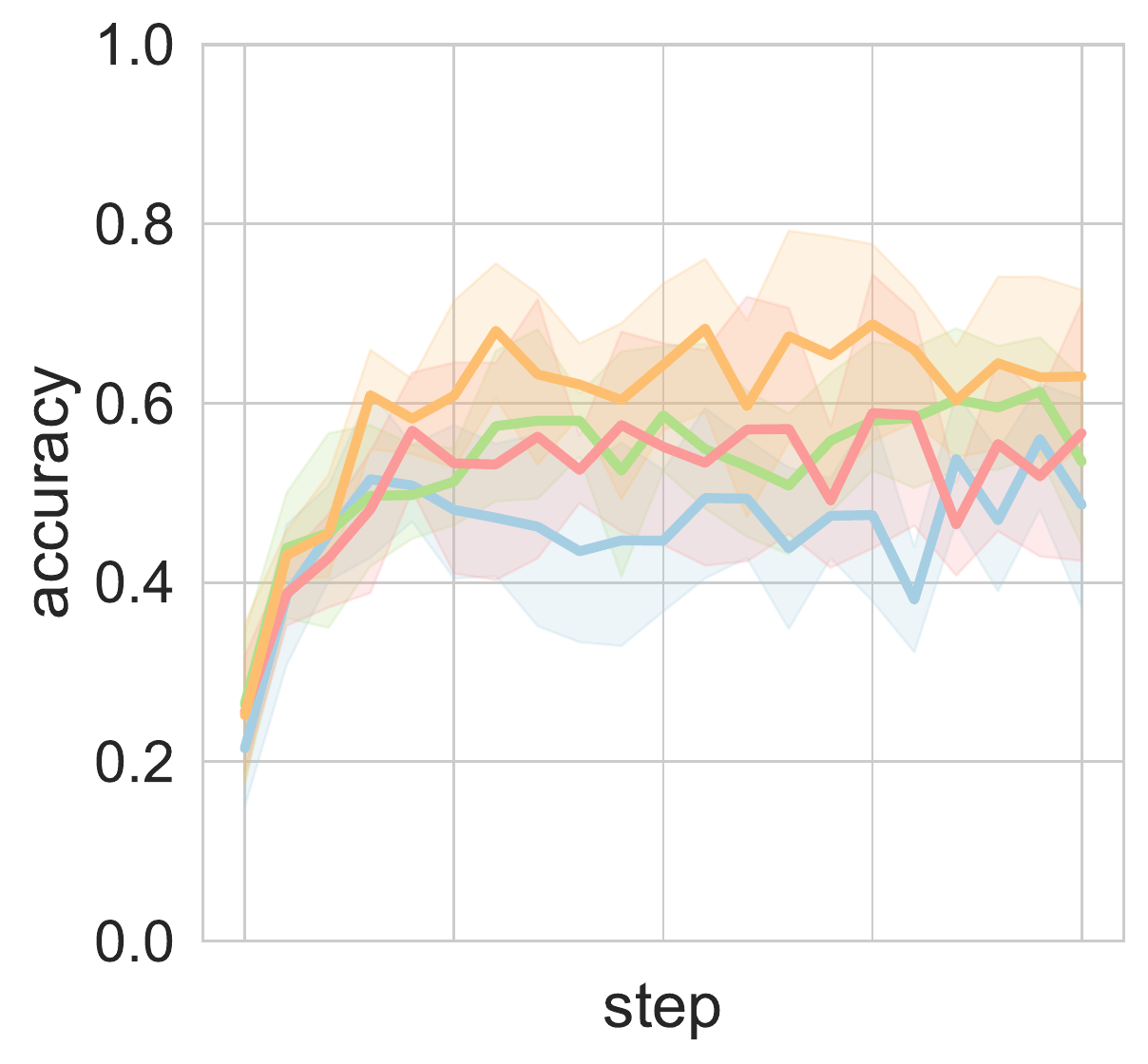}
    \end{minipage}
    \label{fig:and_mask_color_acc}
    }
    \subfigure[Union, digit accuracy.]{
    \begin{minipage}{4cm}
    \includegraphics[width=4cm]{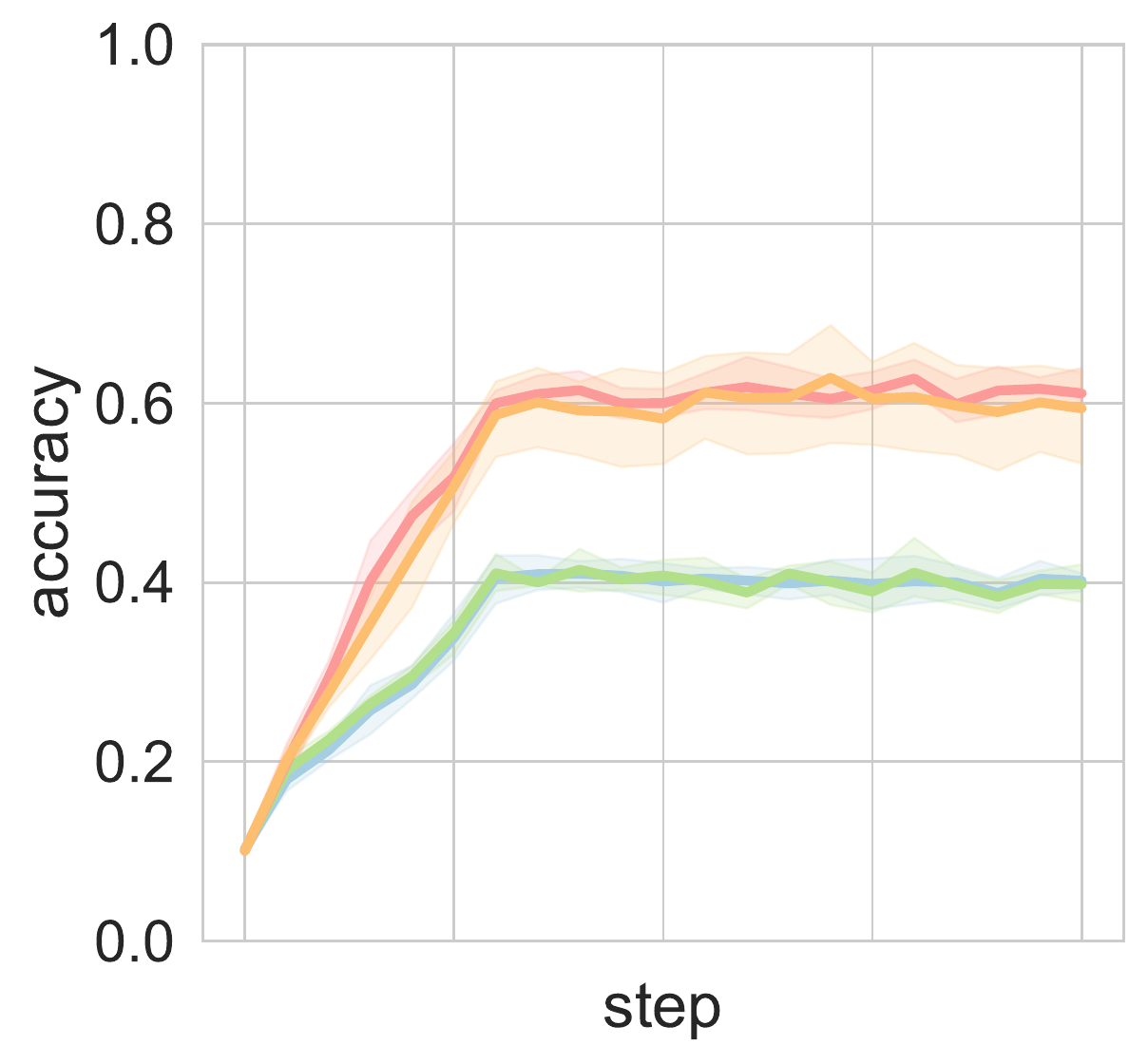}
    \end{minipage}
    \label{fig:or_mask_acc}
    }
    \subfigure[Union, color accuracy.]{
    \begin{minipage}{4cm}
    \includegraphics[width=4cm]{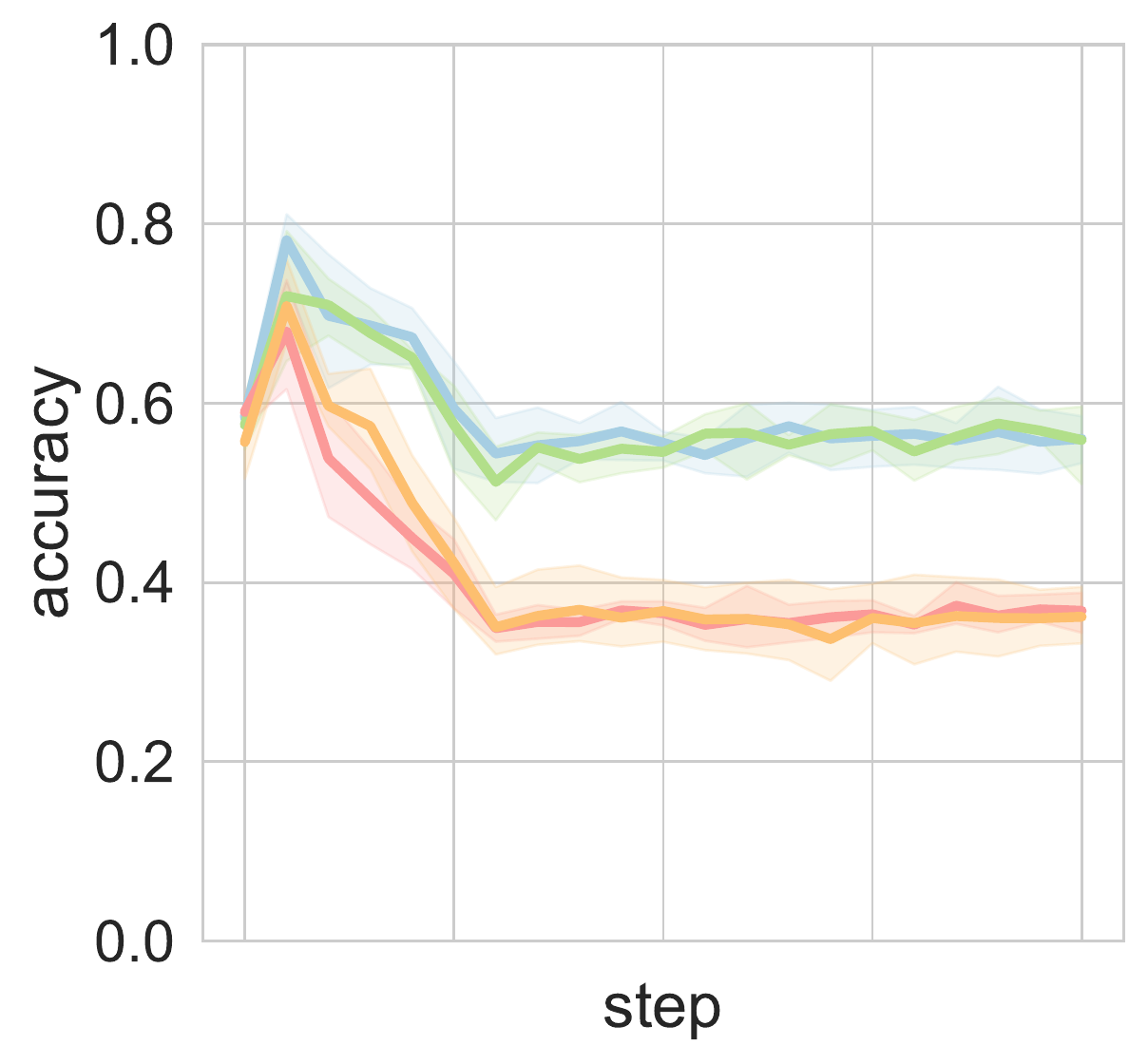}
    \end{minipage}
    \label{fig:or_mask_color_acc}
    }
    \subfigure[Complement of intersection, digit accuracy.]{
    \begin{minipage}{4cm}
    \includegraphics[width=4cm]{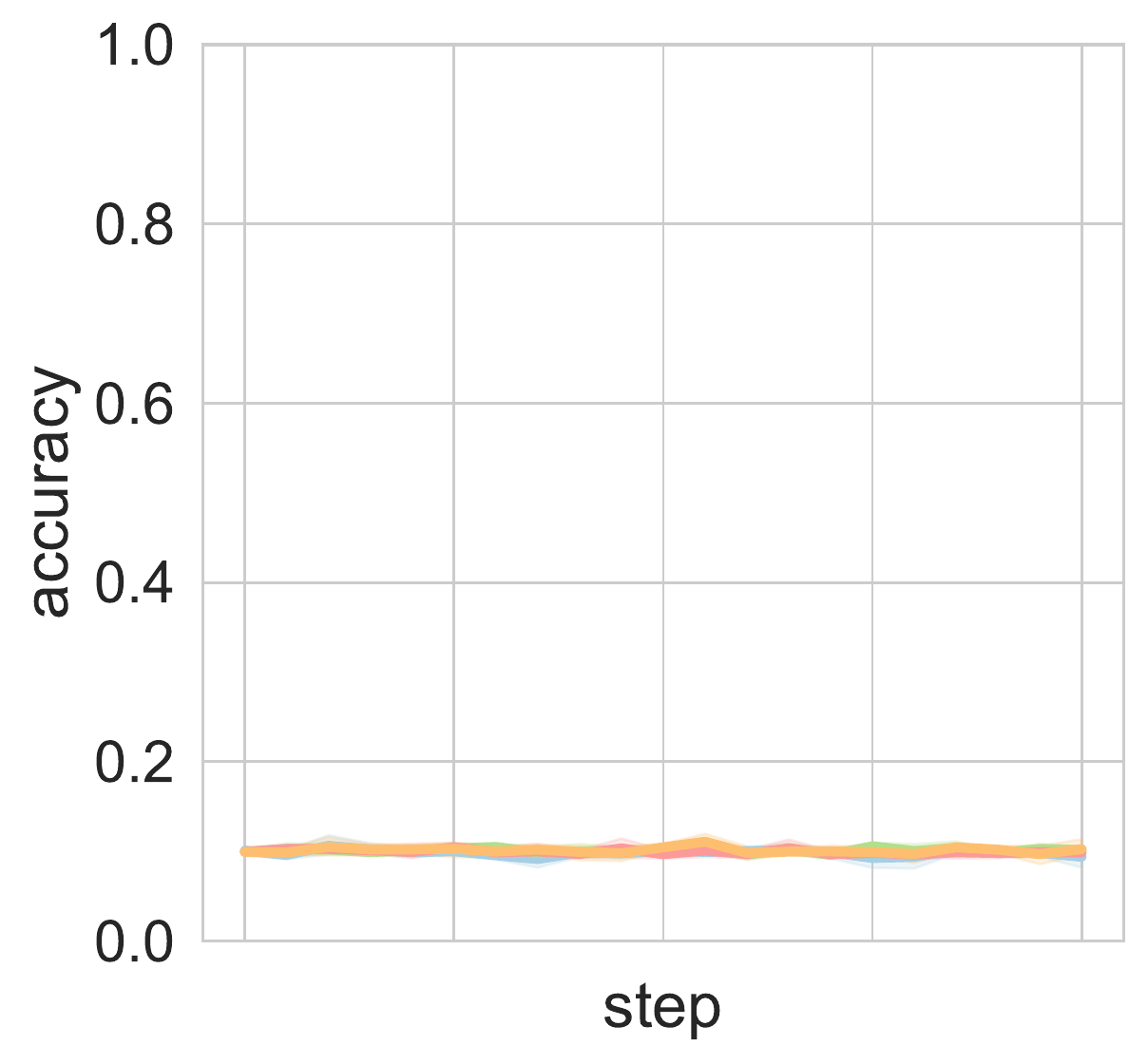}
    \end{minipage}
    \label{fig:not_digit_and_color_acc}
    }
    \subfigure[Complement of intersection, color accuracy.]{
    \begin{minipage}{4cm}
    \includegraphics[width=4cm]{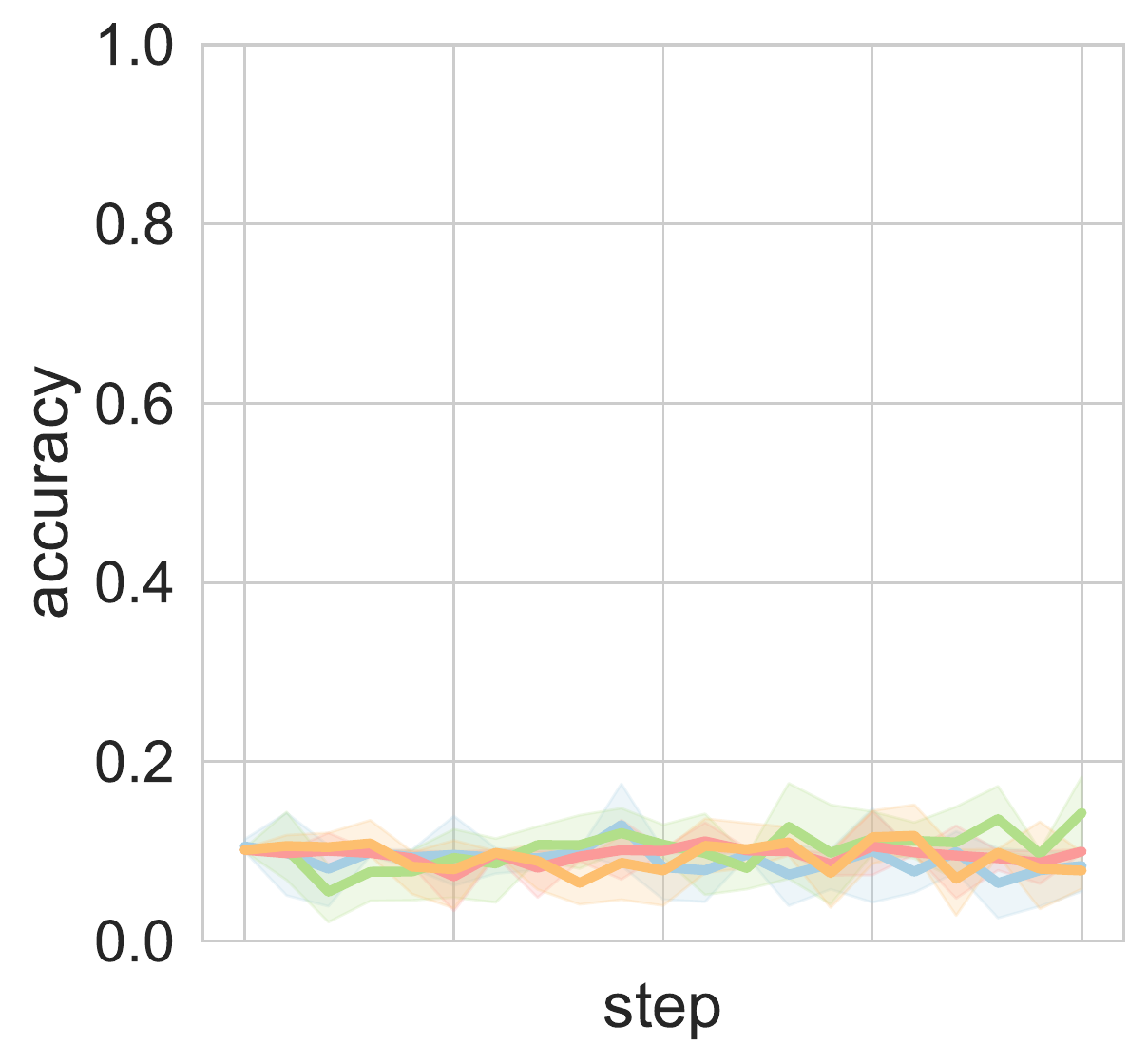}
    \end{minipage}
    \label{fig:not_digit_and_color_color_acc}
    }
    \subfigure[(Complement of color module) $\cup$ (digit module), digit accuracy.]{
    \begin{minipage}{4cm}
    \includegraphics[width=4cm]{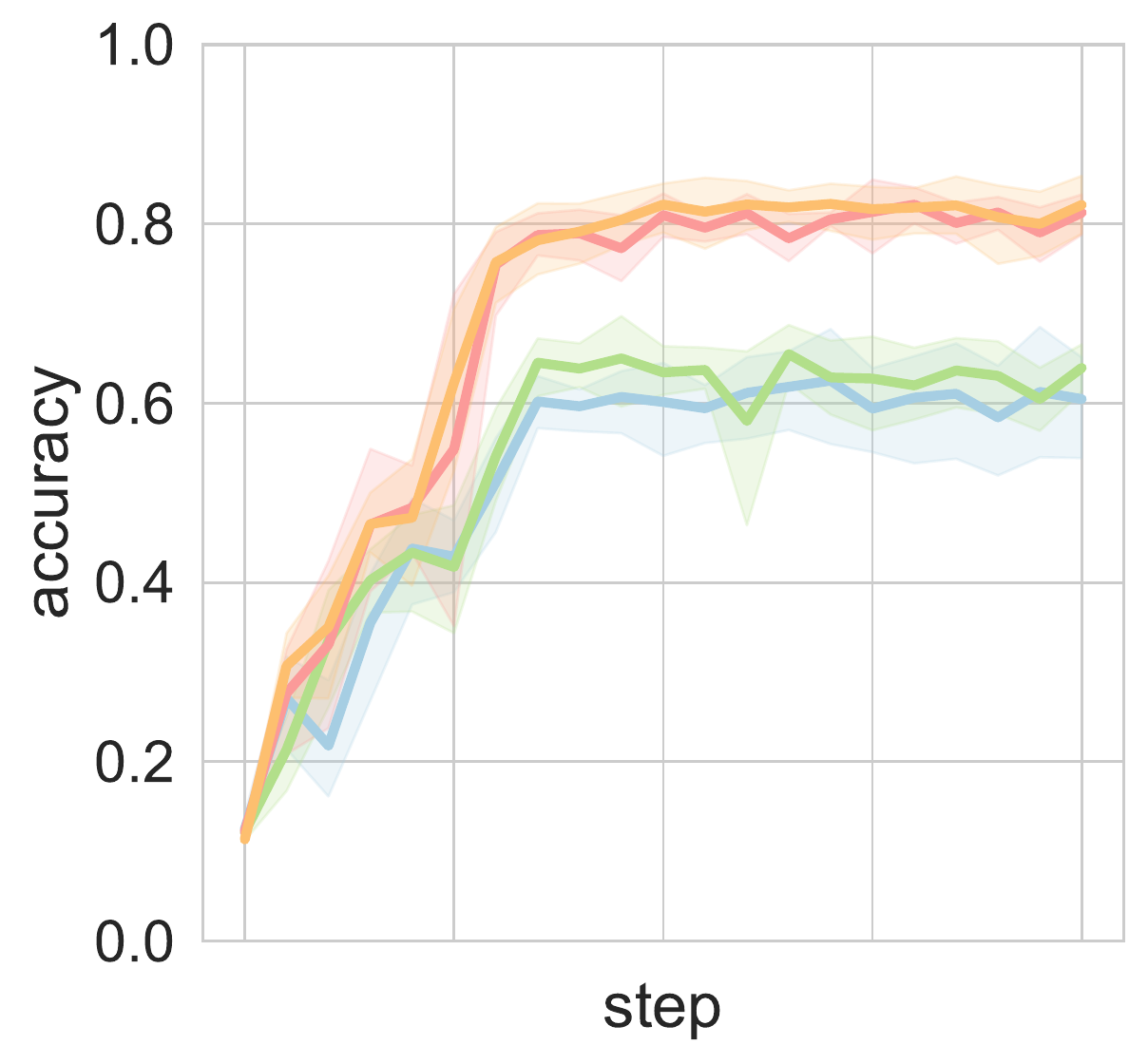}
    \end{minipage}
    \label{fig:not_color_rid_digit_acc}
    }
    \subfigure[(Complement of color module) $\cup$ (digit module), color accuracy.]{
    \begin{minipage}{4cm}
    \includegraphics[width=4cm]{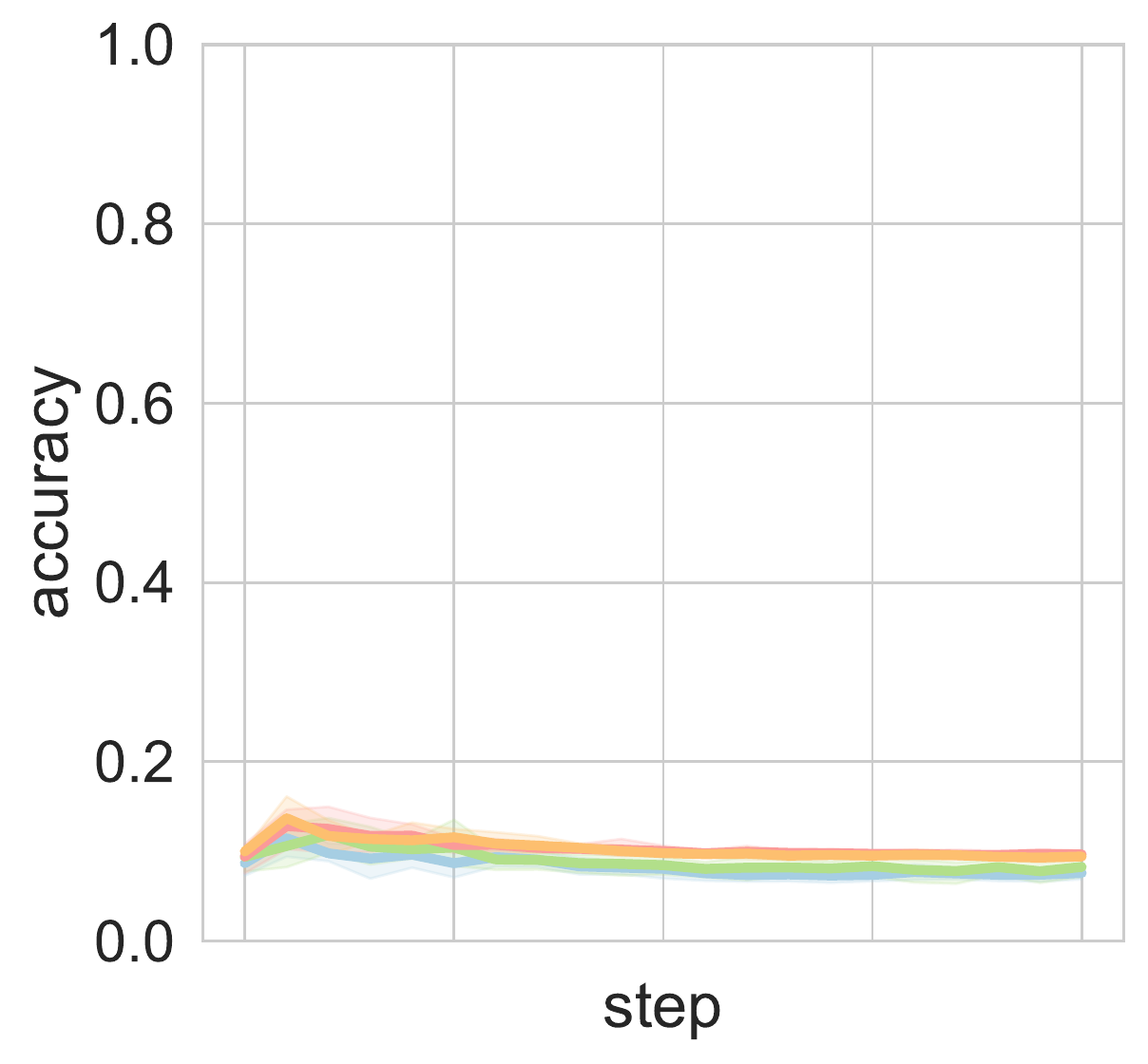}
    \end{minipage}
    \label{fig:not_color_rid_digit_color_acc}
    }
    \caption{The digit / color accuracy for some logical operation results of learned digit and color module.}
    \label{fig:mask_logic_operation}
\end{figure}

\begin{figure}[th]
    \subfigure[Visualization of modules for the linear layer.]{
    \begin{minipage}{12cm}
    \includegraphics[width=12cm]{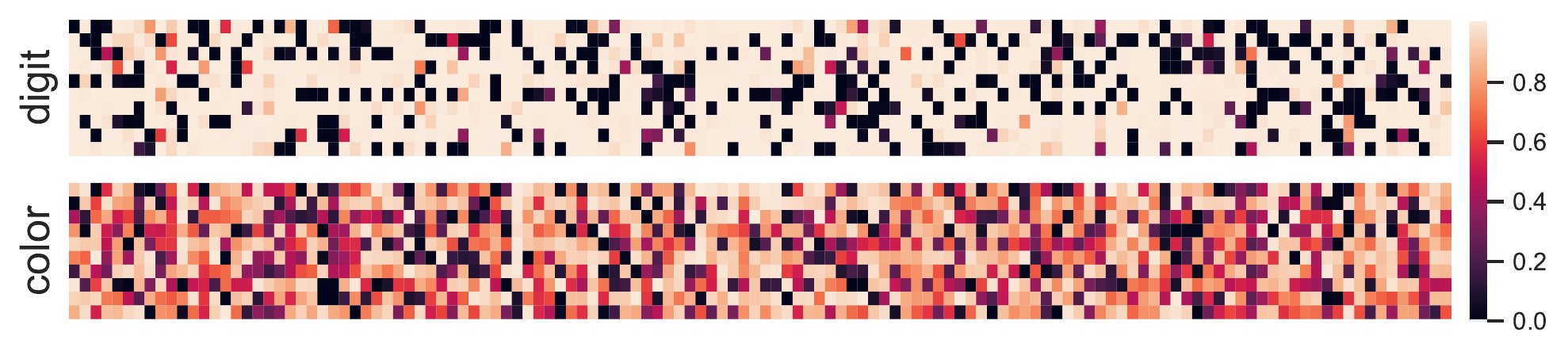}
    \end{minipage}
    \label{fig:visual_fc}
    }
    \subfigure{
    \begin{minipage}{4cm}
    \includegraphics[width=4cm]{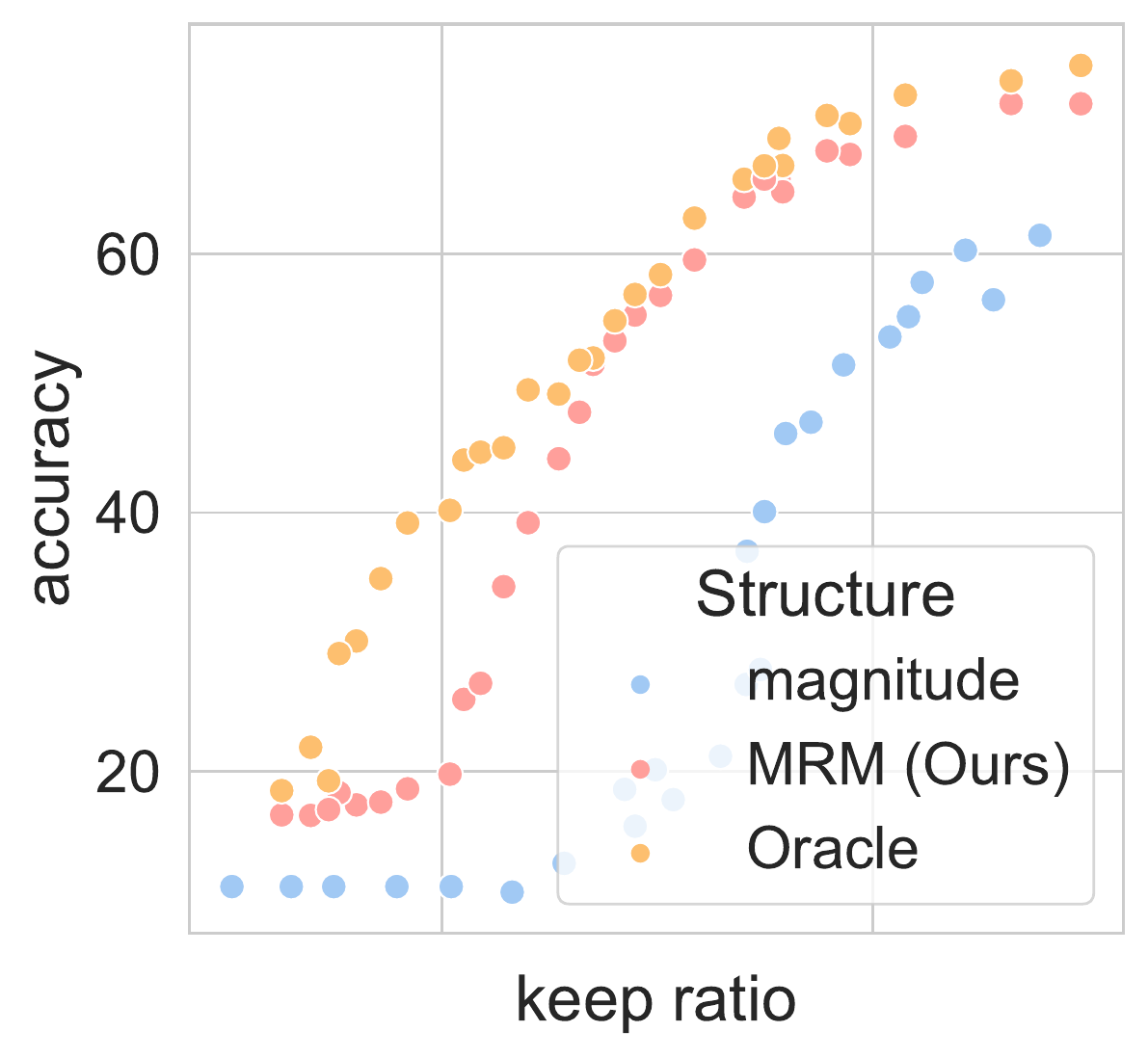}
    \end{minipage}
    \label{fig:mag}
    }
    \caption{Compensatory experimental results.}
\end{figure}

We play with the obtained digit and color module with several basic logical operations in Figure~\ref{fig:mask_logic_operation}.
The intersection of digit and color module is important as its complement behaves trivially in Figure~\ref{fig:not_digit_and_color_acc} and \ref{fig:not_digit_and_color_color_acc}. However, the intersection module alone can not express predictiveness (Figure~\ref{fig:and_mask_acc}) for digit identification. We show that  it needs to be combined with other part to work in Figure~\ref{fig:not_color_rid_digit_acc} and \ref{fig:not_color_rid_digit_color_acc}. We also additionally visualize the two modules for the linear layer in Figure\ref{fig:visual_fc} and get similar results to that in Figure~\ref{fig:visualize}.



For completeness we also compare with magnitude pruning, although it's not for out-of-distribution generalization and its structure search process is data independent like us. The results for ERM on \textsc{FullColoredMNIST} are in Figure~\ref{fig:mag}.
We can see our methods surpass this widely used pruning approach (\textit{e.g.}, by the original lottery ticket hypothesis paper and most prevailing approaches  \cite{han2015deep, zhu2017prune}) in all level of sparsity. Please see more related discussion in Section~\ref{sec:new_pruning}.
\label{sec:mask_ablation}
Furthermore, we do ablation about the hyperparameters of subnetwork learning in the stage 2 of MRM algorithm in Table~\ref{tab:mask_ablation} to show the importance of proper level of ``good" sparsity (instead of random sparsity, whose importance is restricted as discussed in main text).
In Table~\ref{tab:n2} we also show results with different values of $N_2$ in MRM algorithm. Although $N_2$ we adopt is not the optimal setting, we simply use it since it's enough to achieve an approximate convergence of stage 2.
We further do ablation about MRM for with or without stage 3 in Table~\ref{tab:mask_mrm_stage}, showing the importance of the last stage and the validity of our functional lottery ticket hypothesis.

\begin{table}[t]
\caption{Ablation for hyperparameters of subnetwork learning. All numbers are out-distribution generalization accuracy.}
\label{tab:mask_ablation}
\vskip 0.15in
\begin{center}
\begin{sc}
\begin{minipage}[b]{.75\linewidth}
\begin{tabular}{c | ccccc}
\toprule
\diagbox[width=4em]{lr}{$\alpha$} & $1e-3$  & $1e-4$ & $1e-5$ & $1e-6$ & $1e-7$ \\
\midrule 
$1e-2$ & 50.45 \scriptsize{$\pm$ 2.70} & 68.56 \scriptsize{$\pm$ 0.39} & 67.46 \scriptsize{$\pm$ 0.42} &64.13 \scriptsize{$\pm$ 1.05} & 63.25 \scriptsize{$\pm$ 1.12} \\
$1e-1$  & 36.95 \scriptsize{$\pm$ 2.25} & 67.83 \scriptsize{$\pm$ 0.81} & 72.98 \scriptsize{$\pm$ 0.58} & 71.91 \scriptsize{$\pm$ 0.85} & 71.30 \scriptsize{$\pm$ 0.55} \\
$1$ & 21.76 \scriptsize{$\pm$1.29} & 50.19 \scriptsize{$\pm$ 2.33} & 65.54 \scriptsize{$\pm$ 2.16} & 66.31 \scriptsize{$\pm$ 2.56} & 65.51 \scriptsize{$\pm$ 3.31} \\
\bottomrule
\end{tabular}
\end{minipage}
\noindent
\begin{minipage}[b]{.2\linewidth}
\label{tab:n2}
\begin{tabular}{c | cc}
\toprule
$N_2$ & accuracy \\
\midrule
1000 & 71.46 \scriptsize{$\pm$ 1.89} \\
2000 & 72.98 \scriptsize{$\pm$ 0.58} \\
3000 & 73.39 \scriptsize{$\pm$ 0.84} \\
5000 & 73.70 \scriptsize{$\pm$ 0.59} \\ 
\bottomrule
\end{tabular}
\end{minipage}

\end{sc}
\end{center}
\vspace{-0.5cm}
\end{table}


\begin{table}[t]

\caption{Ablation for the third stage of MRM.}
\label{tab:mask_mrm_stage}
\vskip 0.15in
\begin{center}
\begin{sc}
\begin{tabular}{c | cc}
\toprule
Methods & w/ stage 3 & w/o stage 3 \\
\midrule 
MRM & 72.98 \scriptsize{$\pm$ 0.58} & 62.99 \scriptsize{$\pm$ 1.96} \\
ModIRM  & 70.86 \scriptsize{$\pm$ 2.12} & 58.87 \scriptsize{$\pm$ 2.40}  \\
ModREx & 82.06 \scriptsize{$\pm$ 0.73} & 77.48 \scriptsize{$\pm$ 2.30} \\
ModDRO & 85.53 \scriptsize{$\pm$ 0.61} & 80.47 \scriptsize{$\pm$ 1.87} \\
\bottomrule
\end{tabular}
\end{sc}
\end{center}
\end{table}

\subsection{Towards a different pruning method}
\label{sec:new_pruning}
In this paper we propose an OOD algorithm coined as MRM. Here we point out that MRM can actually be seen as a different pruning method and can be applied to IID tasks as well. We do not state this explicitly in the main text since it's not closely related to our OOD research category.
We slightly modify MRM to serve as a pruning algorithm in this way: we also use three stages training paradigm as MRM in main text, but specifically for stage 2, we jointly train the subnetwork structure and model parameters, and let the algorithm break the looping once the specified sparsity level is reached.
We shall not claim this method to be a novel pruning method as we notice that it's similar (though stille different) to \citet{Louizos2018LearningSN}, which for unknown reasons is not taken into baseline consideration by recent pruning works. 
We now show related experimental results in the sense of pruning settings (which is IID generalization) in Table~\ref{tab:cifar_pruning}. The codes and baseline results are based on \citet{wang2020picking}. We also keep the hyperparameters setting to be consistent with \citet{wang2020picking}, and take the modular learning rate $0.01$ and modular sparsity regularization $0.0001$. Our method outperforms other baseline methods, especially for extremely sparse 98\% pruning ratio cases. We think of this as an insightful advantage of our approach.  

\begin{table}[ht]
\caption{Our pruning method on CIFAR10 with ResNet32 across different pruning ratios.}
\label{tab:cifar_pruning}
\vskip 0.15in
\begin{center}
\begin{sc}
\begin{tabular}{c | ccc}
\toprule
\diagbox[width=12em]{Methods}{ratio} & 90\% & 95\% & 98\%\\
\midrule 
Full network & 94.23 & -- & -- \\
\midrule 
OBD \cite{lecun1989optimal} & 94.17 & 93.29 & 90.32 \\
MLPrune \cite{zeng2018mlprune} & 94.21 & 93.02 & 90.31\\
LT \cite{frankle2018lottery} & 92.31 & 91.06 & 88.78 \\
LT rewind \cite{frankle2020pruning} & 93.97 & 92.46 & 89.18 \\
DSR \cite{mostafa2019parameter} & 92.97 & 91.61 & 88.46 \\
SET \cite{mocanu2018scalable} & 92.30  & 90.76 & 88.29 \\
Deep-R \cite{mocanu2018scalable} & 91.62 & 89.84 & 86.45 \\
SNIP \cite{lee2018snip} & 92.59 & 91.01 & 87.51 \\
GraSP \cite{wang2020picking} & 92.38 & 91.39 & 88.81 \\
\midrule
Ours  & 94.19 &  93.36 & 92.80 \\
\bottomrule
\end{tabular}
\end{sc}
\end{center}
\end{table}

\end{document}

%% file: cmd.tex
\usepackage{amsmath,amsfonts,bm,amssymb,mathtools,amsthm}
\usepackage{color,xcolor}
\usepackage{xspace} 
\def\onedot{$\mathsurround0pt\ldotp$}

\newtheorem{proposition}{Proposition}

\def\eqref#1{equation~\ref{#1}}

\def\1{\bm{1}}

\def\vv{{\bm{v}}}

\DeclareMathAlphabet{\mathsfit}{\encodingdefault}{\sfdefault}{m}{sl}
\SetMathAlphabet{\mathsfit}{bold}{\encodingdefault}{\sfdefault}{bx}{n}

\newcommand{\R}{\mathbb{R}}

